\documentclass[letterpaper, 10 pt, conference]{ieeeconf}

\IEEEoverridecommandlockouts
\makeatletter
\let\NAT@parse\undefined
\makeatother

\usepackage{graphics}
\usepackage{graphicx}
\usepackage{grffile}
\usepackage{amsmath}
\usepackage{amssymb}
\usepackage{color}
\usepackage{bm}
\usepackage{url}

\usepackage[noend]{algpseudocode}
\usepackage{algorithm}
\usepackage{makecell}
\usepackage[square, comma, numbers,sort&compress]{natbib}
\usepackage{multirow}
\let\labelindent\relax
\usepackage{enumitem}
\usepackage[table,dvipsnames]{xcolor}
\usepackage[pagebackref=true,breaklinks=true,colorlinks,bookmarks=false]{hyperref}
\usepackage{sidecap}
\usepackage{threeparttable}

\hypersetup{
linkcolor=BrickRed
,citecolor=Green
,filecolor=Mulberry
,urlcolor=NavyBlue
,menucolor=BrickRed
,runcolor=Mulberry
,linkbordercolor=BrickRed
,citebordercolor=Green
,filebordercolor=Mulberry
,urlbordercolor=NavyBlue
,menubordercolor=BrickRed
,runbordercolor=Mulberry
}

\title{\LARGE \bf
A Deep Reinforcement Learning Environment for\\
Particle Robot Navigation and Object Manipulation
}

\author{\small Jeremy Shen$^{1*}$, Erdong Xiao$^{2*}$, Yuchen Liu$^{2}$, Chen Feng$^{2\dagger}$\\
\url{https://ai4ce.github.io/DeepParticleRobot}
\thanks{$^{*}$ indicate equal contributions.}%
\thanks{$^{1}$ Stuyvesant High School, New York, NY 10280, USA
	{\tt\small jshen20@stuy.edu}}%
\thanks{$^{2}$ New York University, Brooklyn, NY 11201, USA
	{\tt\small \{ex276, yl5680, cfeng\}@nyu.edu}}%
\thanks{$^{\dagger}$ Corresponding author. {\tt\small cfeng@nyu.edu}}%
}

\begin{document}

\maketitle
\thispagestyle{empty}
\pagestyle{empty}

\begin{abstract}
Particle robots are novel biologically-inspired robotic systems where locomotion can be achieved collectively and robustly, but not independently. While its control is currently limited to a hand-crafted policy for basic locomotion tasks, such a multi-robot system could be potentially controlled via Deep Reinforcement Learning (DRL) for different tasks more efficiently. However, the particle robot system presents a new set of challenges for DRL differing from existing swarm robotics systems: the low degrees of freedom of each robot and the increased necessity of coordination between robots. We present a 2D particle robot simulator using the OpenAI Gym interface and Pymunk as the physics engine, and introduce new tasks and challenges to research the underexplored applications of DRL in the particle robot system. Moreover, we use Stable-baselines3 to provide a set of benchmarks for the tasks. Current baseline DRL algorithms show signs of achieving the tasks but are yet unable to reach the performance of the hand-crafted policy. Further development of DRL algorithms is necessary in order to accomplish the proposed tasks.
\end{abstract}

\section{Introduction}
Innovations in hardware and computing have sparked increased interest and research in swarm robotics. Swarm robot systems provide a more adaptable, robust, and scalable alternative to current state-of-the-art robots. While traditional swarm robots, like drones, maintain a basic level of capabilities in each individual, a new class of swarm robots with simpler hardware has been proposed: the particle robot. Existing work in particle robots can be classified into two categories: biological-inspired particle robots, like viruses and sea urchins, that can operate independently~\cite{mateos2020particle}, and statistical-mechanics-inspired particle robots that are incapable of self-locomotion~\cite{shuguang2019paricle}. Both types of particle robots pose a collaborative control task, but in this paper, we will focus on the statistical mechanics-inspired particle robot. This type of robot is unable to complete any task on its own, so in order to utilize the particle robot, we must control an extensive multi-robot system.

\begin{figure}[t]
    \centering
    \includegraphics[scale=0.3]{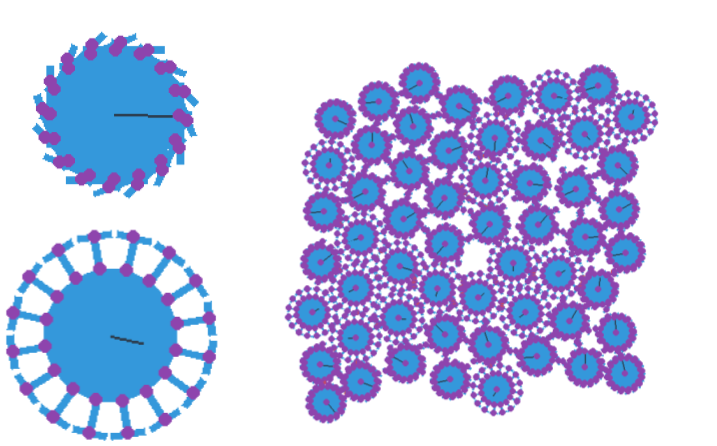}
    \caption{\textbf{Upper left}: Contracted single particle robot. \textbf{Lower left}: Expanded single particle robot. \textbf{Right}: 56-particle robot system with mixed states}
    \label{fig:demo}
\end{figure}

\begin{figure}[t]
    \centering
    \includegraphics[scale=0.3]{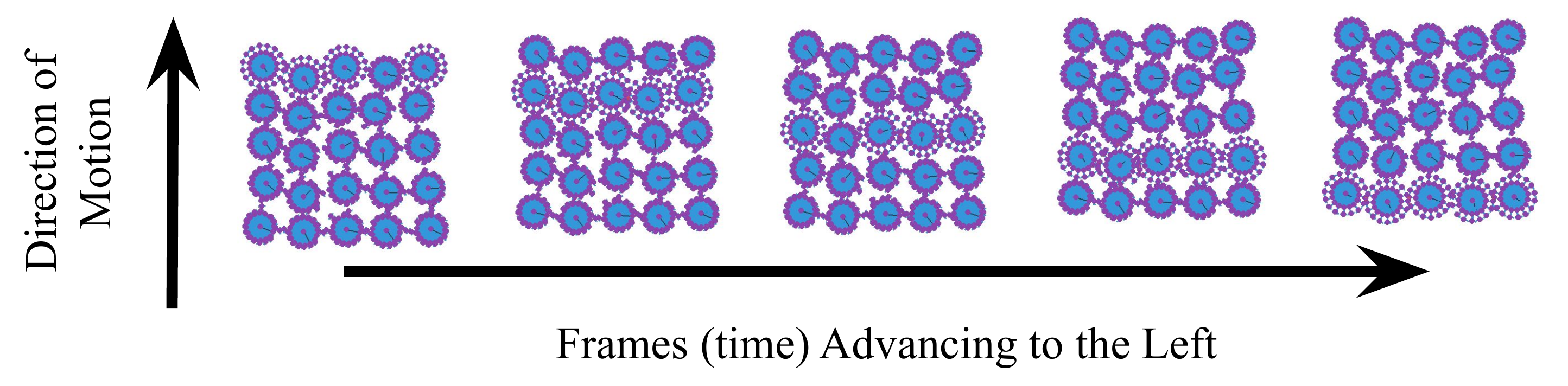}
    \caption{Longitudinal wave policy on a 25-particle robot system. The direction of the expansion-contraction wave is downwards, and the motion of the particle robots is upwards. Each image is an individual frame.}
    \label{fig:samplewavepolicy}
\end{figure}

We present a Particle Robot simulator to study reinforcement learning methods on this type of robots. The particle robot~\cite{shuguang2019paricle} we study in this simulator is a disk-shaped robot only capable of changing its radius (increasing/decreasing its size). It switches between two states: fully expanded (largest size) or fully contracted (smallest size), by moving paddles attached to its central body (see Figure~\ref{fig:demo}). These paddles are not in contact with the ground meaning individual particles are incapable of self-propulsion. However, as a part of a particle robot system, the collective particle robots can leverage the internal forces from the paddle movement between particle robots and the external force from the friction from the ground acting on each particle robot's central body. It has been proven that if the particle robot system's expansion-contraction cycle follows a longitudinal wave, the particle robots' motion will be in the opposite direction to the wave (see Figure~\ref{fig:samplewavepolicy}). For further detail on the dynamics of the particle robot system, please check section~\ref{sec:rob-design}.


Control methods for the particle robot can be mathematically derived, like the aforementioned “wave-policy”, but it would be a difficult and tedious task to create a new control method for each new robot design and task. The appeal of swarm robots is their adaptability, so it makes sense for particle robot control to be easily repurposed and reapplied. This motivates the application of deep reinforcement learning as a control method for particle robots. Since the introduction of Deep Q-Networks (DQN)~\cite{volodymyr2019dqn}, new Deep Reinforcement Learning (DRL) algorithms have become widespread, including Advantage Actor-Critic (A2C)~\cite{mnih2016a2c} and Proximal Policy Optimization (PPO)~\cite{schulman2017ppo}. The combination of traditional Q-Learning or Policy Gradient methods and the powerful modeling ability of deep neural networks has improved the performance of DRL algorithms, allowing them to tackle increasingly difficult problems. DRL has become well suited for complex robot control problems, thus it is reasonable to study if DRL methods could automatically find optimal control policies for the particle robot system. Our particle robot simulator poses a novel swarm robot environment with unique constraints on individual robot capabilities to test and develop DRL methods.

As the first step of studying DRL for particle robots, we choose to use a centralized control method to facilitate collaborative particle robots. We consider this a ``super-agent'' setup, where there exists an outside controller that possesses full observability of the world and the ability to send simultaneous instructions to every particle robot. Although the particle robot system seems naturally more suited for decentralized RL, and the super-agent setup seems to make the problem easier due to the full observability and coordination, we find our setup already non-trivial and challenging for existing DRL baselines. Note that our environment supports the decentralized setup, but in this work, we focus on thoroughly studying the super-agent control before advancing to a more challenging problem.

Our simulator contains tasks varying in difficulty from point-to-point navigation to object manipulation, and allows for varying the number and states of the particle robots. By using the OpenAI gym interface, it is possible to modify observation and action spaces, tasks and challenges, and reward methods, and it is easy to apply reinforcement learning techniques on the particle robots.

We also benchmark four baseline algorithms: (1) a hand-crafted control policy, (2) DQN, (3) A2C, and (4) PPO, over four different tasks: (1) simple navigation, (2) obstacle navigation, (3) navigation with unresponsive particles, and (4) object manipulation.

In summary, our contributions are the following:
\begin{itemize}

    \item introduced a new deep reinforcement learning environment for particle robot research.
    \item implemented multiple, challenging tasks for the development of particle robots.
    \item benchmarked three existing deep reinforcement learning algorithms on our tasks.

\end{itemize}

\section{Related Work}
\textbf{Particle-robot-like Swarm Robot Control}.
There are existing low-capability robot systems similar to particle robots which are not controlled using deep reinforcement learning. Research is focused on exploiting the randomness of collective low-level actions to generate movement in larger robot systems. The Smarticle~\cite{savoie2019smarticle} system implements statistical models and unsupervised learning to create asymmetries in smarticle collisions by activating and deactivating certain smarticles. Smarticle control was advanced with a more physics-inspired model for self organization~\cite{chvykov2021rattling}. Deblais et al.~\cite{deblais2018boundaries} considers a bounded collection of rod-like robots which can form mobile and deformable superstructures. Boudet et al.~\cite{boudet2021scaffold} proposes a Langevin’s equation-inspired dynamics model for this robot scaffold. These systems are collections of robots bounded by a soft or hard container, creating parallels to fluid behaviors. The particle robots in our environment are not container bounded but use the magnetic attraction for a similar adhesive purpose as in~\cite{shuguang2019paricle}. 

Wang et al.~\cite{wang2020distribute} propose the Virtual Particle Exchange, a localization algorithm that relies on simple sensors and calculations and can anonymize and limit robot communications. Mayya et al.~\cite{8967985} considers equilibrium thermodynamics in brushbot systems to take advantage of the differences in robot densities. Yang et al.~\cite{8884178} creates vortex-like paramagnetic nanoparticle swarms to track and control microbots. Like particle robots, individual robots in these systems possess little to no self-locomotion abilities. Unlike previous works that handcraft control policies from physics rules, with our particle robot environment, we aim to use deep reinforcement learning to discover explicit control policies. The robots used in our simulator are inspired by particle robots~\cite{shuguang2019paricle}, a disk-shaped, biological-cell-inspired robot only capable of changing its size and without self-locomotion abilities. Researchers hand-derived a sinusoidal-function control policy (an expansion/contraction cycle determined from distance to goal position) for large swarms of particle robots to complete various navigation tasks.

\textbf{Deep Reinforcement Learning-based Control Method}.
Deep reinforcement learning has become popular for solving multi-agent system control problems. Collision avoidance between robots is a critical problem for multi-robot systems; Fan et al.~\cite{fan2018collision} and Long et al.~\cite{long2018collision} have developed reinforcement learning based control methods for multi-robot navigation tasks by combining a collision penalty term and the time cost term in their reward function, enabling the robots to learn a time-optimal and collision-free trajectory. DRL is capable of coordinating systems of robots that possess significantly more locomotive abilities than our particle robots. Drone system control is a more difficult multi-agent control problem which presents a larger state and action space since drones can move freely in 3D space, but deep reinforcement learning is still viable for cooperative multi-drones tasks. Real-time trajectory planning of multi-drone systems~\cite{rldrone2016arya} trains a large amount of drones (with limited communication) using a PPO algorithm to find a collision-free trajectory. This demonstrates the PPO algorithm's capability of solving cooperative navigation problems, which is similar to our proposed Particle Robot tasks. Wang et al.~\cite{wang2019pattern} introduces an application of DRL in pattern formation: a multi-robot system learns to configure in given patterns such as lines, circles, and more complex shapes; these tasks could show the potential of using DRL to solve similar cooperation tasks for particle robot systems. Moreover, Ding et al.~\cite{ding2020distributed} shows that robots can learn a cooperative strategy in the multi-arms manipulation task. Zhu et al.~\cite{zhu2020flocking} discusses using DRL to solve the robot flocking control problem (a migrating-birds-like robots system). Sun et al.~\cite{sun2021motion} give a systemic review of DRL in the robot motion planning field. DRL has proven to be a successful method of discovering robot control policies, and it is considered to have the potential to solve the particle robots' control problem.

\textbf{Related Reinforcement Learning Environments}.
It is necessary for our environment to preserve realism and maximize adaptability so that the control methods developed in the simulator could also work with real-world particle robots. Thus, our Deep Particle Robot simulator (DPR) features a continuous world environment, centralized and decentralized communication structures, navigation and manipulation tasks, and a physics engine (PhysE). Our environment can be easily modified to add new robot designs, challenges, and goals, and the Pymunk physics engines simulates realistic forces and collisions. Compared to our environment, there are various existing environments for multi-agent reinforcement learning, but none of them are directly suitable for simulating particle robots (see Table~\ref{tab:sim-envs}). Multi-Agent-Reinforcement-Learning-Environment~\cite{jiang2018} offers a framework for multi-agent problems like playing soccer, navigating mazes, and manipulating boxes. MAgent~\cite{zheng2017magent} is one of the first environments to support millions of agents and offer numerous variants of communication structures and tasks. Multi-agent-Particle-Envs~\cite{Lowe2020maparicle} has a physics engine collision handler and both a continuous and discretized action space. However, these environments are simplified to a discretized state space and non-continuous world, limiting simulator realism. Multi-Agent-Emergence-Environments~\cite{Baker2020maEmergent} is solely a hide-and-seek game and every agent within a team shares its observations. Playground~\cite{playground}, an environment for the multi-agent game Pommerman, emphasizes coordination and strategy but uses a grid world and a binary action space. There exists specialized simulators for video games like Neural-MMOs~\cite{neural-mmo}, SMAC~\cite{starcraft} for Starcraft II, and Google Research Football~\cite{football} for soccer, and while these simulators are used for multi-agent reinforcement learning research, they are unable to be applied in particle robots.

\begin{table}[H]
\centering
\caption{Related Multi-agent Reinforcement Learning Environments}\label{tab:sim-envs}
\begin{threeparttable}

\begin{tabular}{|c|c|c|c|c|c|c|c|}\hline
&State\tnote{a}&Comm\tnote{b}&PhysE\tnote{c}&Nav\tnote{d}&Manip\tnote{e}&\#Agents\tnote{f}&PR\tnote{g}\\
\hline
\cite{jiang2018}&D&C&$\times$&\checkmark&\checkmark&$<$10&$\times$\\
\hline
\cite{zheng2017magent}&D&D&$\times$&\checkmark&$\times$&$>$100&$\times$\\
\hline
\cite{Lowe2020maparicle}&D&D&2D&\checkmark&$\times$&10-100&$\times$\\
\hline
\cite{Baker2020maEmergent}&C&C&3D&\checkmark&\checkmark&$<$10&$\times$\\
\hline
\cite{playground}&D&D&$\times$&$\times$&$\times$&$<$10&$\times$\\
\hline
\cite{neural-mmo}&C&C+D&$\times$&$\times$&\checkmark&$<$100&$\times$\\
\hline
\cite{starcraft}&C&D&$\times$&\checkmark&$\times$&$<$100&$\times$\\
\hline
\cite{football}&C&C+D&$\times$&$\times$&\checkmark&10-100&$\times$\\
\hline
Ours&C&C+D&2D&\checkmark&\checkmark&$<$100&\checkmark\\
\hline
\end{tabular}
\begin{tablenotes}
\item[a] Continuous state (C) or discretized state/grid world (D)
\item[b] Centralized (C) or decentralized (D) communication structure
\item[c] 2D, 3D, or No Physics Engine
\item[d] Navigation Task
\item[e] Manipulation Task
\item[f] Number of Agents Supported
\item[g] Low-level locomotion-inhibited robotics environment
\end{tablenotes}
\end{threeparttable}
\end{table}

\section{Environment Setup}
Our Particle Robots environment uses Pymunk, a 2D Physics library, to simulate the environment, and Pygame for visualization. The environment also uses the OpenAI Gym interface, allowing for easy reinforcement learning integration.
\subsection{Robot Design}\label{sec:rob-design}
\begin{figure}[htp]
    \centering
    \includegraphics[scale=0.2]{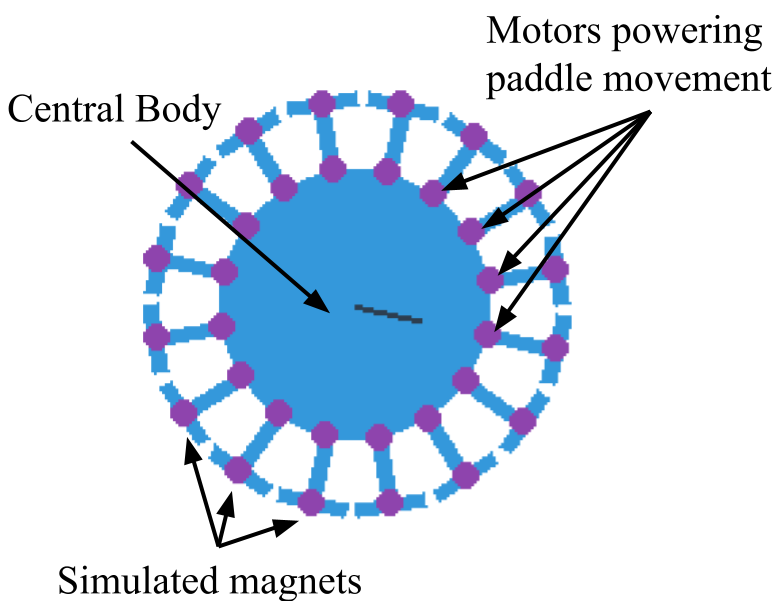}
    \caption{Anatomy of Particle Robot}
    \label{fig:singlepr}
\end{figure}

The basic Particle Robot consists of a circular central body with sixteen paddles attached to it. Each paddle has a range of 90 degrees, from being tangent to being perpendicular to the central body, and is powered with a simple motor. A second set of paddles is attached to the end of the first set of paddles which rotate to remain parallel to the surface of the central body (see Figure \ref{fig:singlepr}).

The paddles of each particle robot are lined with magnets in order to exert an attractive force on its neighboring particle robots. The magnetic force is necessary to keep the particle robot system together.

The original paper of Particle Robot~\cite{shuguang2019paricle} has shown that the force from the paddles, friction force, and magnetic force all must lie within a specific range but has not provided definite values:
\begin{equation}
F_{f, i} < F_{paddles/magnetic} < \sum_{j = 1}^{n} F_{f, j}, \quad \forall i,
\label{eq:sysdynamics}
\end{equation}
where \(F_{f, i}\) is the magnitude of the friction force on the $i$-th particle robot, \(F_{paddles}\) is the force the paddles exert, and \(F_{magnetic}\) is the magnetic force.
    
\subsection{Observations}
Observations are comprised of any combination of information derived from a top-down camera. The top-down camera is centered over the goal, and the position, velocity (linear and rotational), distance to goal, and state of each particle robot and any objects can be captured. It is also possible to directly use each Pygame frame as the observation. 

\subsection{Actions}
	\textbf{Individual robot}. The basic particle robot has two states, fully contracted and fully expanded. Actions cannot be overridden, so when the particle robot's paddles are in motion, all proposed actions will be voided. This discretized action space is used for our experiments, but it is possible to use a continuous action space to set an exact angle for the paddle's state, allowing the particle robot to partially expand.
    
	\textbf{Super-agent}. Our super-agent setup treats the individual particle robots within its system as another degree of freedom. Each particle robot is assigned an index, and the action space can be a list of binary states (multi-binary action space) or a decimal integer (discrete action space) to be converted into a binary value. Each particle robot will take its corresponding action based on its index.
	
\subsection{Reward Design Motivation}
New reward functions can easily be added to the Gym interface. The reward functions are the same for any type of agent: the center of mass of the Particle Robot system, individual particle robots, or manipulated objects. 

We present a piecewise reward function that is used in our experiments where \(\mathbf{G}\) is the coordinate of the goal, \(\mathbf{A}_t\) is the coordinate of the Agent's position at time \(t\), and \(\theta\) is the angle between \(\mathbf{G} - \mathbf{A}_{t-1}\) and \(\mathbf{A}_t - \mathbf{A}_{t-1}\):
\begin{equation}
r_{t} = \begin{cases}
||\mathbf{A}_t - \mathbf{A}_{t-1}||  \frac {d\cos(\theta)} {||\mathbf{G} - \mathbf{A}_t||}, \quad & ||\mathbf{G} - \mathbf{A}_t|| \leq d;\\
||\mathbf{A}_t - \mathbf{A}_{t-1}|| \cos(\theta), \quad & ||\mathbf{G} - \mathbf{A}_t|| > d. \\
\end{cases}
\label{eq:reward}
\end{equation} 

Note that \(||\mathbf{A}_t - \mathbf{A}_{t-1}|| \cos(\theta)\) is the length of the agent's displacement vector (previous position to current position) projected onto the ideal vector (previous position to goal position). This is the distance the agent travels in the correct direction. When the agent is within a threshold distance \(d\), the magnitude of the reward is scaled inversely proportional to the distance. This helps the model converge as the agent finalizes its approach to the goal.

\begin{figure}[t]
    \centering
    \includegraphics[scale=0.5]{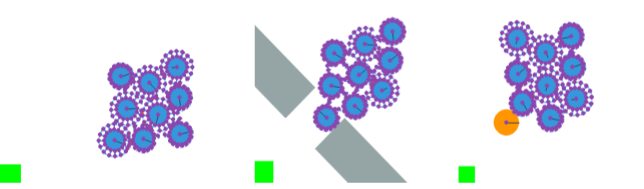}
    \caption{\textbf{Left} Simple navigation task. \textbf{Middle} Obstacle navigation task: gray blocks are the gates. \textbf{Right} Object manipulation task: orange ball is the target object. The goal for all three tasks is the green square in the lower left}
    \label{fig:tasks}
\end{figure}

\subsection{Navigation Task}
\subsubsection{Problem formulation}\label{sec:naviprobformulation} For an N-particle robot system. The state space at time step t, where \(0 < t < T = 2,500\):
\begin{equation}
s_t = (\boldsymbol{x}_{t}, \boldsymbol{y}_{t}, \boldsymbol{\dot{x}}_{t}, \boldsymbol{\dot{y}}_{t}),  \label{eq: navi_state}
\end{equation} 
where $\boldsymbol{x}_{t}$ and $\boldsymbol{y}_{t}$ are the x-coordinates vector and y-coordinates vector respectively of the particle robots.

The action space is N-dimensional binary, where each particle robot can only contract (0) or expand (1) and the action of the entire system is the joint action of all individual particle robots.

The reward \(r_{t}\) takes on the same form as equation~\eqref{eq:reward}, where the agent is the center of mass of the entire particle robot system.
The score function is:

\begin{equation}
 J(\theta) = \sum_{k=0}^{T-1} r_{k+1} \gamma^{T-k-1},
\label{eq:score}
\end{equation}
and our goal is to find a policy function that maximizes the score:

\begin{equation}
 \pi(s;\theta) = \mathop{\arg\max}_{\theta}J(\theta).
 \label{eq:policy}
\end{equation}

\subsubsection{Navigation Variations}
\textbf{Simple navigation}. This is the plainest task designed to demonstrate the Particle Robots ability to perform basic locomotion. The Particle Robots are initialized in a grid at a set start position, and the task is complete when the center of mass of the Particle Robot system reaches the goal position (see Figure \ref{fig:tasks}).
   
 \textbf{Obstacle navigation}. The obstacle navigation task test the Particle Robots' ability to respond to external forces. A gate is set up between the start and goal locations such that the opening is greater than the width of one Particle Robot and less than the width of the entire Particle Robot system (see Figure \ref{fig:tasks}). The shape of the particle robot system will be deformed when going through the gate.
  
\textbf{Navigation with unresponsive particle robots}. The navigation task can be made more challenging by "killing" a fixed number of Particle Robots. Any action sent to an unresponsive Particle Robot is ignored, and the unresponsive Particle Robot will remain in a fully contracted position. Unresponsive Particle Robots still have magnetic attractions, meaning these particle robots will travel with the system. This demonstrates the particle robot system's ability to continue operating despite some faulty robots.    
  
\subsection{Object Manipulation Task:} 
\subsubsection{Problem formulation}\label{sec:manipprobformulation} For an N-particle robot system with a target object. At time step t, where \(0 < t < T = 2,500\):
\begin{equation}
s_t = (\boldsymbol{x}_{t}, x_{t}^{o}, \boldsymbol{y}_{t}, y_{t}^{o}, \boldsymbol{\dot{x}}_{t}, \dot{x}_{t}^{o}, \boldsymbol{\dot{y}}_{t}, \dot{y}_{t}^{o}).  
\label{eq: manip_state}
\end{equation} 

Note that this state space is nearly identical to the state space introduced in equation~\eqref{eq: navi_state}, except that the position and velocity of the target object are also included, denoted with the superscript ``o''.
The action space of the object manipulation task is identical to that of the navigation tasks.

The reward \(r_{t}\) takes on the same form as equation~\eqref{eq:reward}, where the agent is the manipulated object.

The score function for the object manipulation task is identical to that of the navigation tasks (reference equation~\eqref{eq:score}), and our goal is to find a policy function of the same form as equation~\eqref{eq:policy}.

There is significant overlap between the problem formulations of the navigation tasks (section \ref{sec:naviprobformulation}) and the object manipulation task (section \ref{sec:manipprobformulation}). The key difference is that the reward of the navigation tasks is dependent on the particle robots, while the reward of the manipulation task is solely dependent on the manipulated object.

\subsubsection{Manipulation Task}
The object manipulation task tests the Particle Robots ability to apply an external force on a non-Particle Robot object (see Figure \ref{fig:tasks}). The Particle Robots are required to move an object from a start position to a goal position. The shape, dimensions, and size of this object can be changed to simulate different real world objects (ball, brick, door, etc.). To simplify the problem, the object is initialized adjacent to and in between the particle robot system and the goal location. The difficulty of this task is that the Particle Robots must learn and maintain the contact point between the system and the object.

\subsection{Single-agent vs. Multi-agent Environment Setup }
The proposed tasks remain identical between the single and multi-agent setups, but the control methods for the Particle Robots change. 

The single-agent environment regards the $N$ agents system as a super-agent with $N$ degrees of freedom, therefore the action space is the union of the action space for a single agent in the multi-agent setup. This assumption could simplify the problem as a single agent reinforcement learning, and there are already many existing baseline algorithms to help solve the problem. However, this assumption also naturally limits the potential of the multi robots system, for example, the number of particle robots must remain constant.

There are many possible configurations for the multi-agent environment. For our experiments, we use a network architecture where the observation and reward is shared among all the particle robots. It is possible to configure the environment with other structures such as a fully decentralized, competitive setting.

\section{Experiments}
\subsection{Baseline Methods}
We benchmarked the following four baseline methods for particle robot control.

\textit{Hand-crafted algorithm:} We implemented the wave policy discussed in section \ref{sec:rob-design} and shown in equation~\eqref{eq:sysdynamics} and figure~\ref{fig:samplewavepolicy}. The wave policy function returns a sequence of actions that simulates one full expansion/contraction cycle.

\textit{DQN, A2C, PPO:} We used stable-baselines3\cite{stable-baselines3} to implement a discrete setting for DQN, A2C, and PPO.

\subsection{Evaluation Metrics}
Each algorithm is given a fixed number of steps to run (2500), and we evaluate a model's performance given the following results.

\textit{Total distance traveled}: Compared with the net displacement, the total distance the particle robot system travels gives us information about how efficient the system's motion is.

\textit{Net Displacement}: The total displacement of the particle robot system (distance between initial and final coordinates) gives us insight into the system's locomotive abilities. This does not take into consideration the direction of motion.

\textit{Projected Displacement}: The displacement vector is projected onto the ideal trajectory (vector pointing from start to goal). This is the distance the system moves in the correct direction.
\begin{equation}
Projection = ||\boldsymbol{A}_f - \boldsymbol{A}_0|| \cos(\theta)
\label{eq:projection_formula}
\end{equation} 
Where, in equation~\eqref{eq:projection_formula}, \(\boldsymbol{A}_t\) is the agent's position at step \(t \in [0, f]\) and \(\theta\) is the angle between the agent's displacement vector and the ideal trajectory.

\subsection{Significant Hyperparameter Motivations}
\textbf{Learning rate}. The initial learning rate must be large to take advantage of the early exploration steps but the learning rate must eventually decrease in order to converge to an optimal policy which resulted in us using a linear schedule learning rate (from 0.0075 to 0).

\textbf{Network architecture}. After experimenting with different architectures, we discovered that a Q-network/Policy network architecture of [64, 128, 128, 128, 64] (4 hidden layers) resulted in the quickest convergence to an optimal policy.

\textbf{Training length}. We chose 10,000,000 to be our total timesteps during training. This ensures that by the end of the training, the value loss and policy gradient loss (if it exists) converges to 0. 
\subsection{Results}
\begin{figure}[h]
     \centering
     \begin{tabular}{@{}c@{}}
         \includegraphics[scale=0.25]{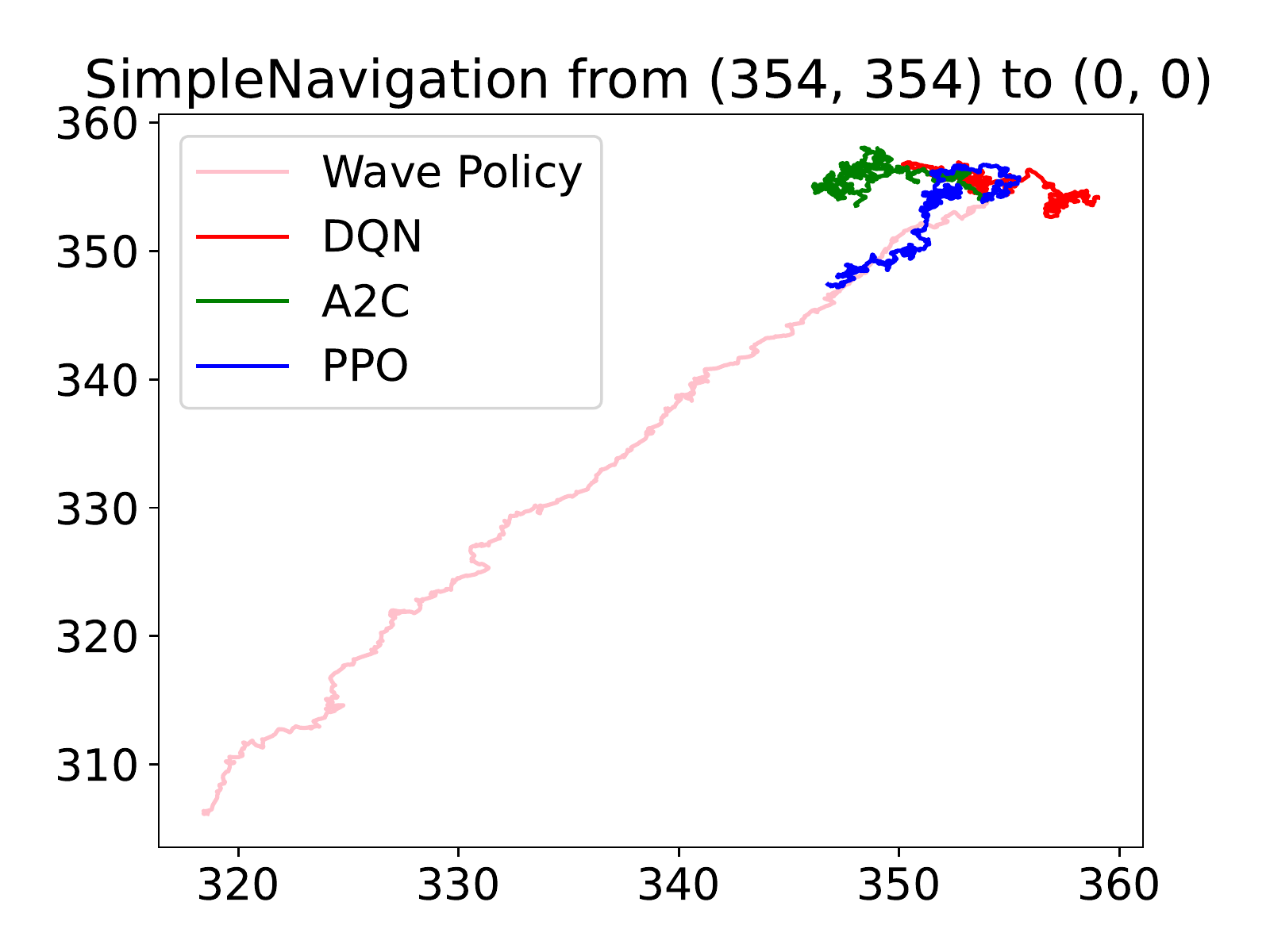}\\[\abovecaptionskip]
         \label{SNp2p}
     \end{tabular}
     \begin{tabular}{@{}c@{}}
         \includegraphics[scale=0.25]{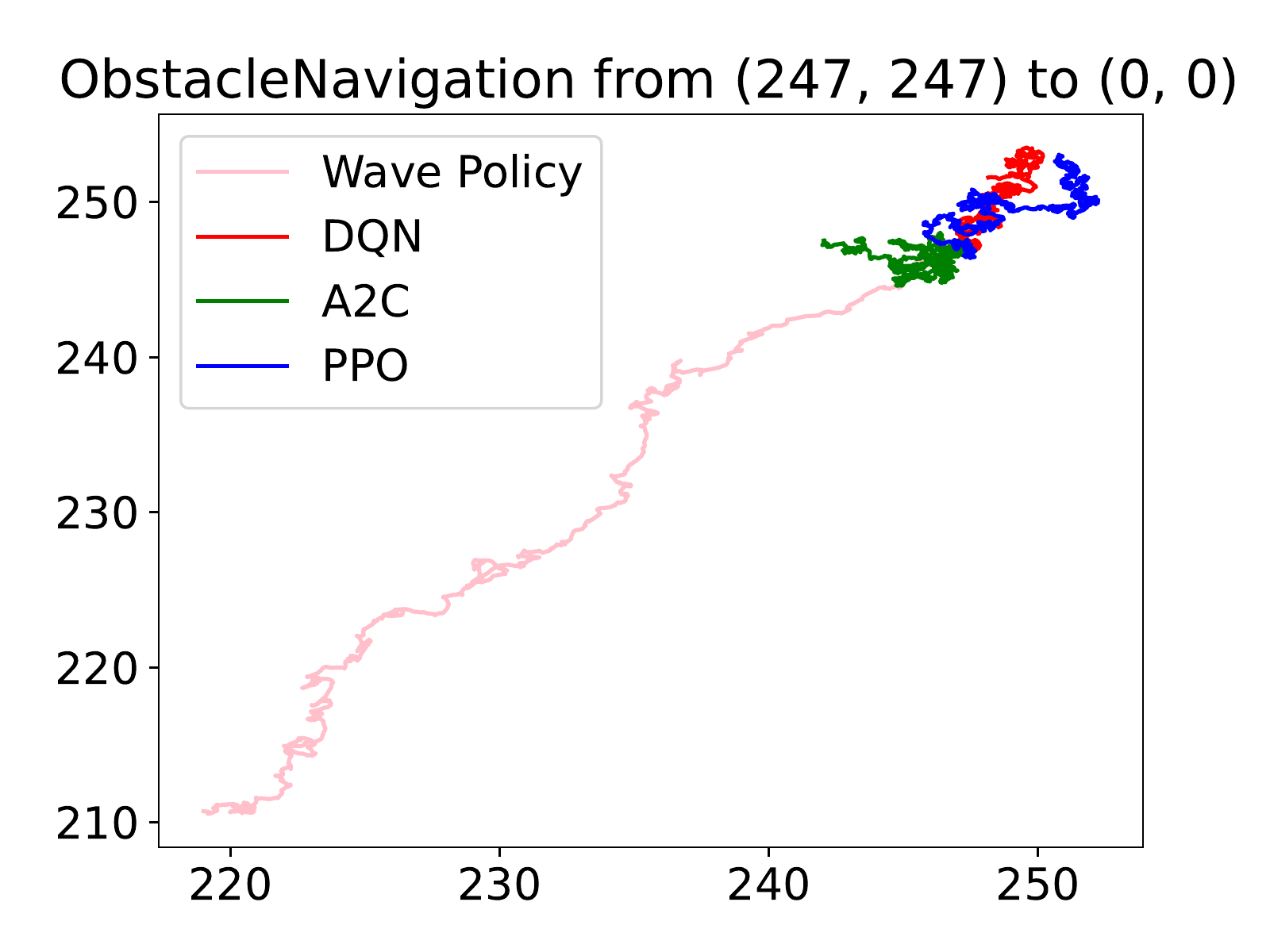}\\[\abovecaptionskip]
         \label{ONp2p}
     \end{tabular}
     
     \begin{tabular}{@{}c@{}}
         \includegraphics[scale=0.25]{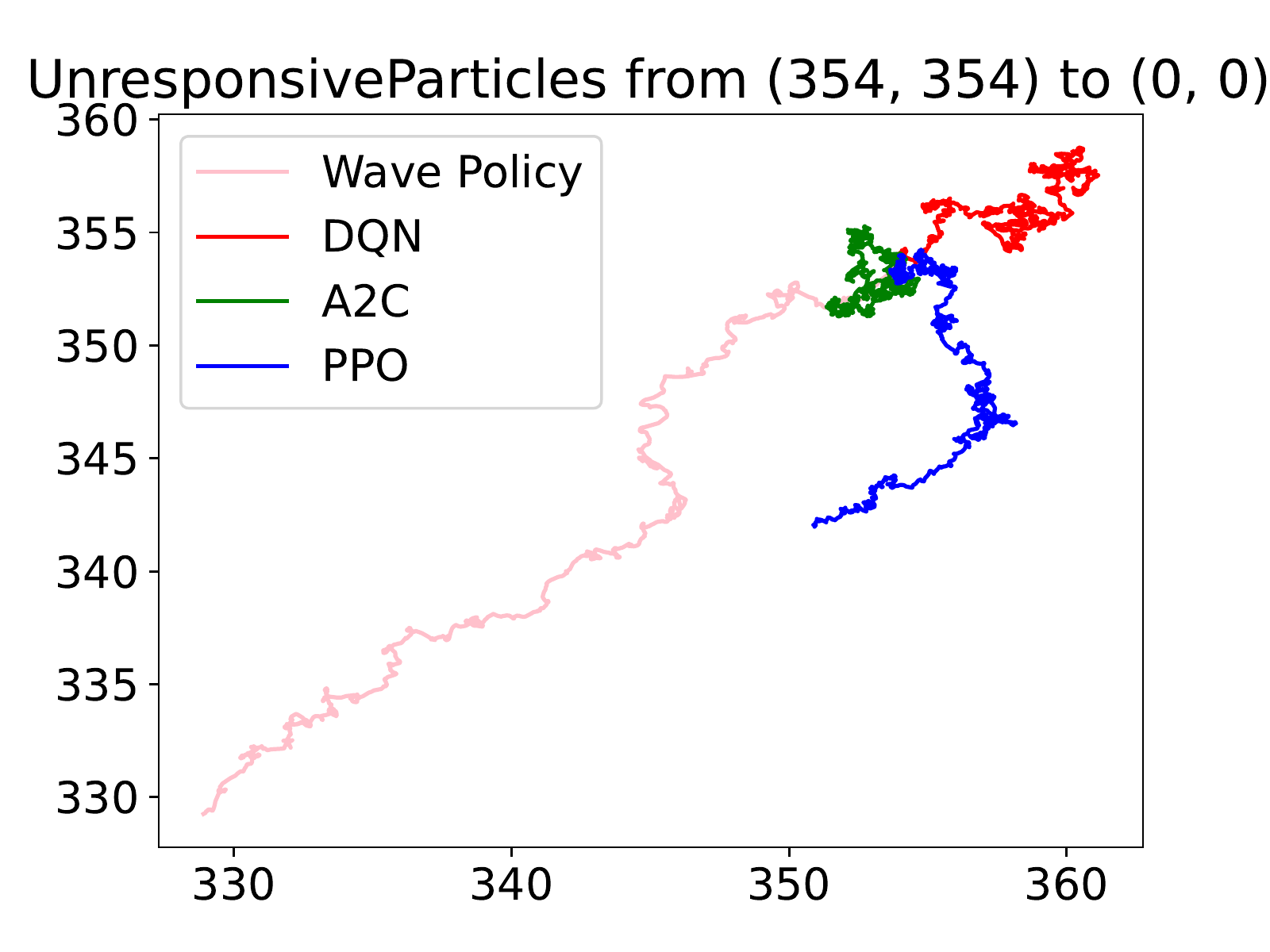}\\[\abovecaptionskip]
         \label{UPp2p}
     \end{tabular}
     \begin{tabular}{@{}c@{}}
         \includegraphics[scale=0.25]{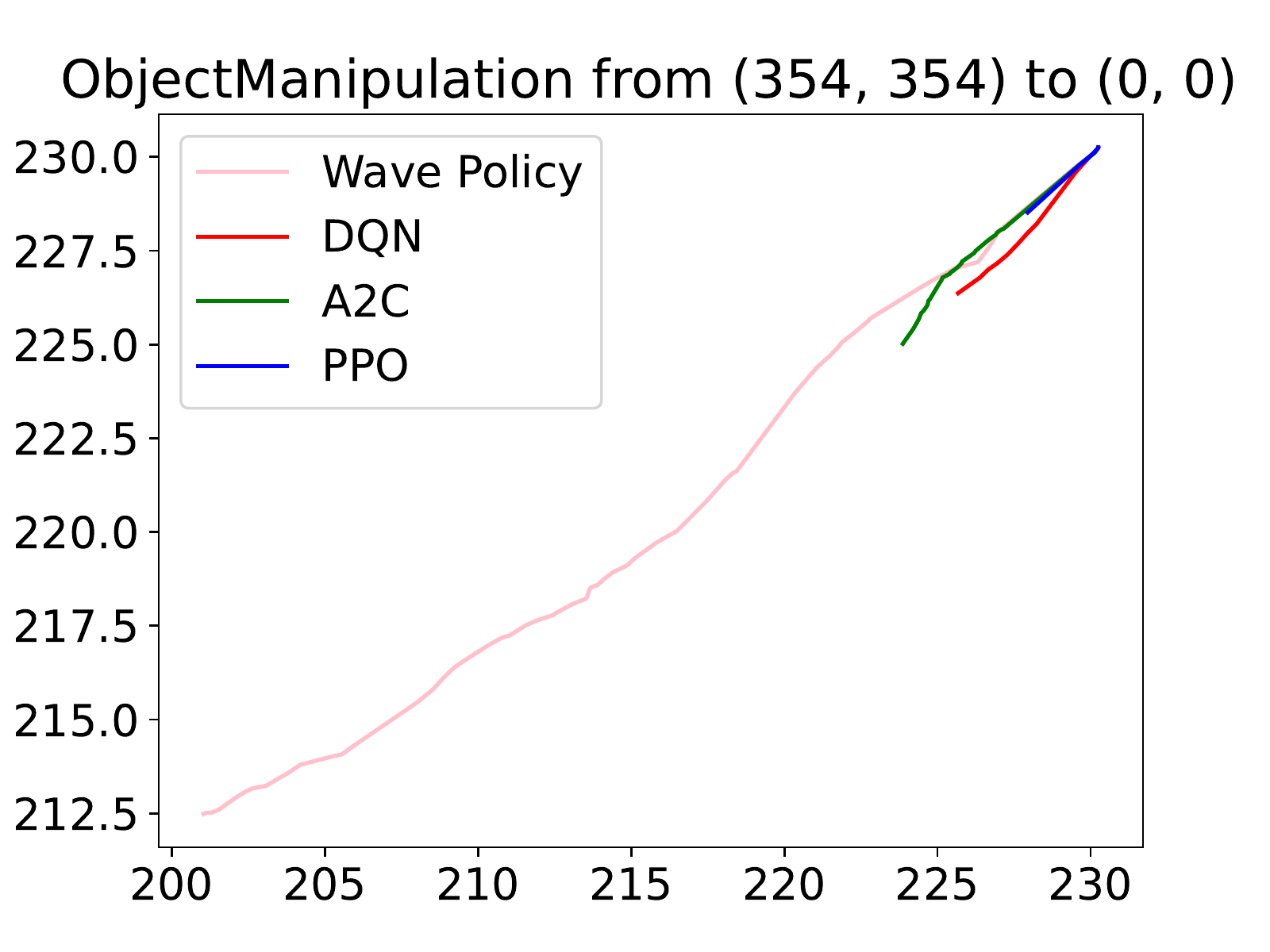}\\[\abovecaptionskip]
         \label{OMp2p}
     \end{tabular}
        \caption{Sample trajectories for (starting upper-left, clockwise) Simple Navigation, Obstacle Navigation, Unresponsive Particles, and Object Manipulation tasks. Each figure simulates the trajectory of the four baseline models all starting from the same point.}
        \label{fig:p2p trajectory}
\end{figure}

\begin{figure*}[ht]
     \centering
     \begin{tabular}{@{}c@{}c@{}c@{}}
         \includegraphics[scale=0.25]{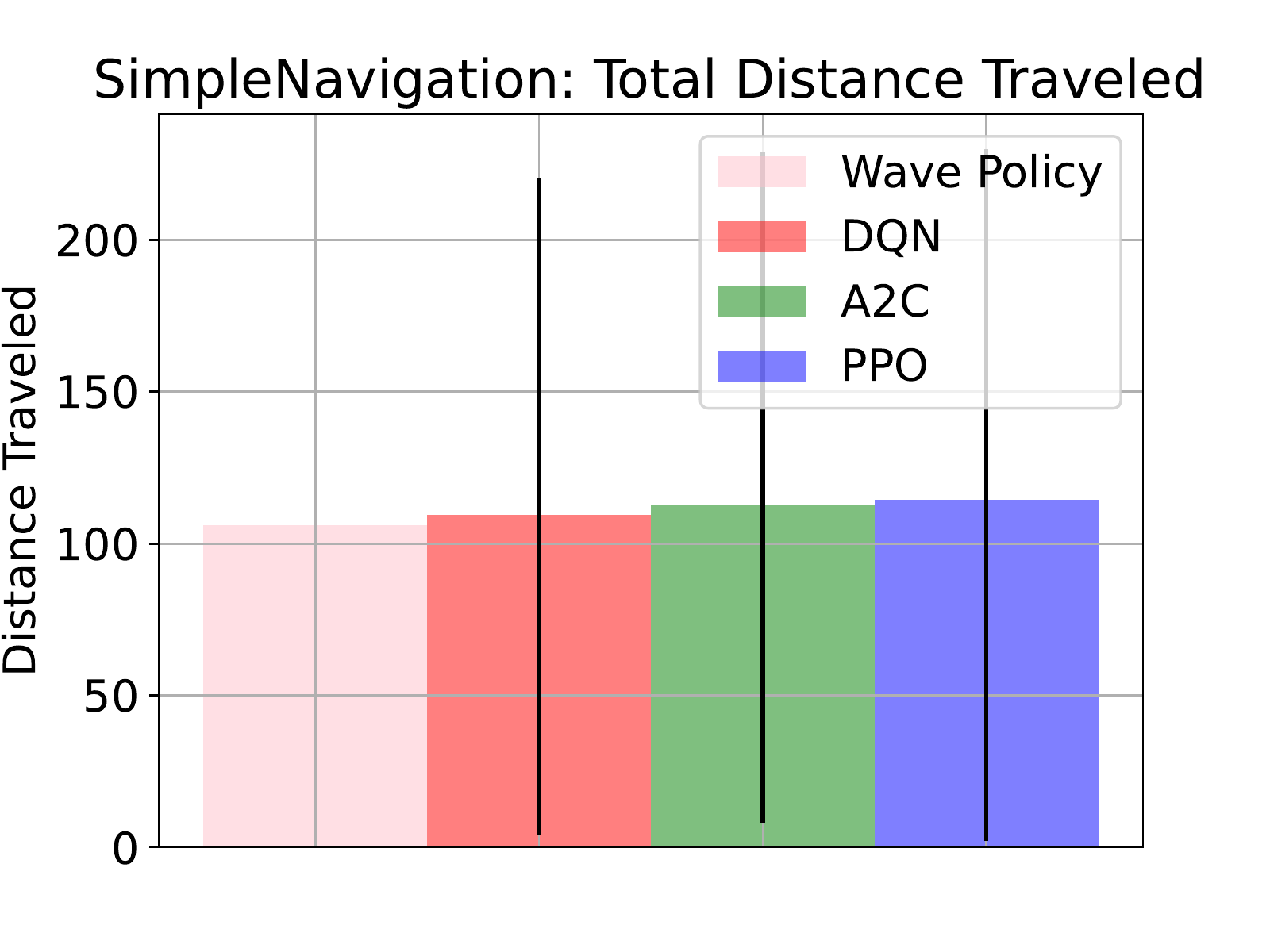}\\[\abovecaptionskip]
     \end{tabular}
     \begin{tabular}{@{}c@{}c@{}c@{}}
         \includegraphics[scale=0.25]{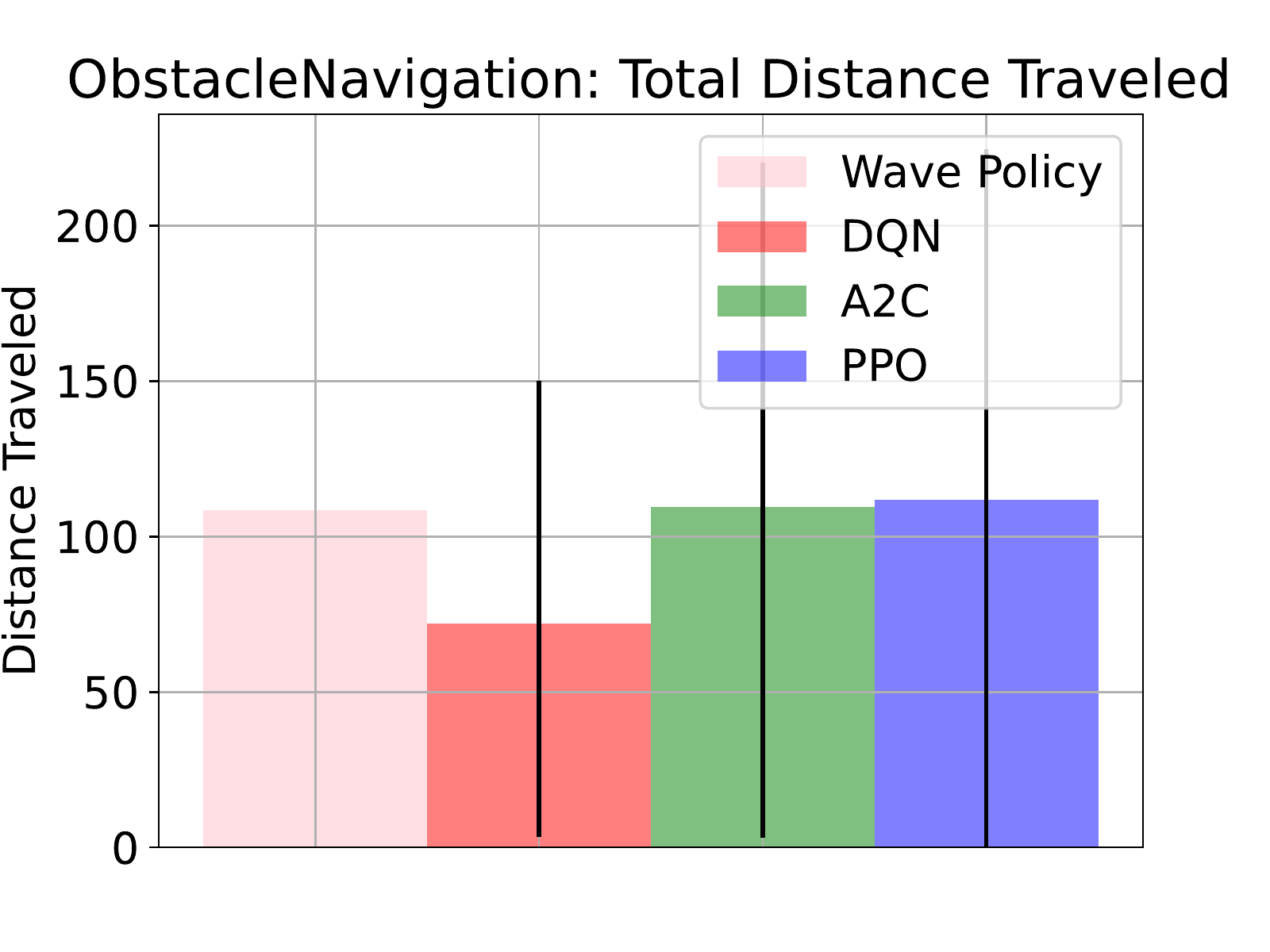}\\[\abovecaptionskip]
     \end{tabular}
     \begin{tabular}{@{}c@{}c@{}c@{}}
         \includegraphics[scale=0.25]{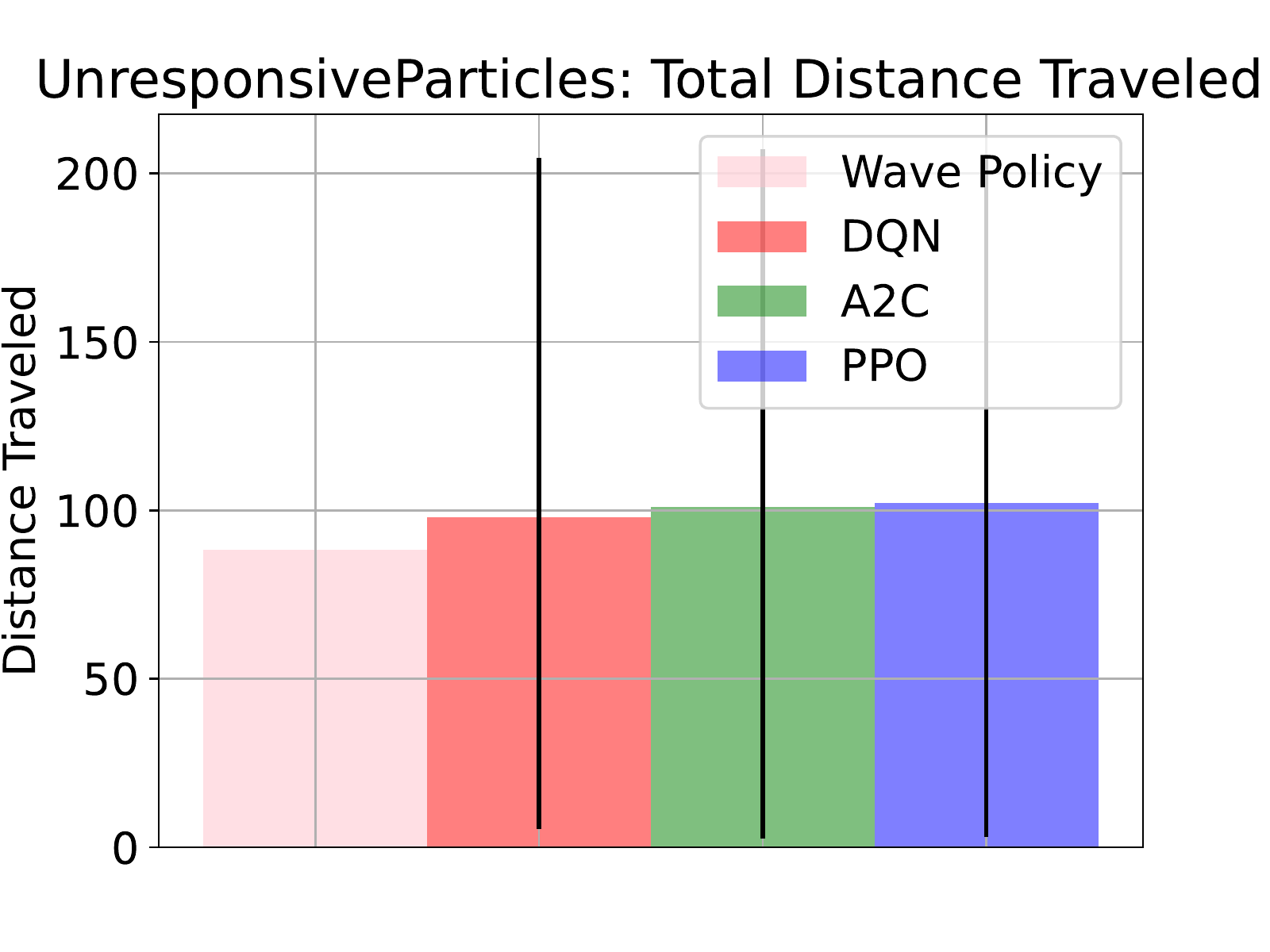}\\[\abovecaptionskip]
     \end{tabular}
     \begin{tabular}{@{}c@{}c@{}c@{}}
         \includegraphics[scale=0.25]{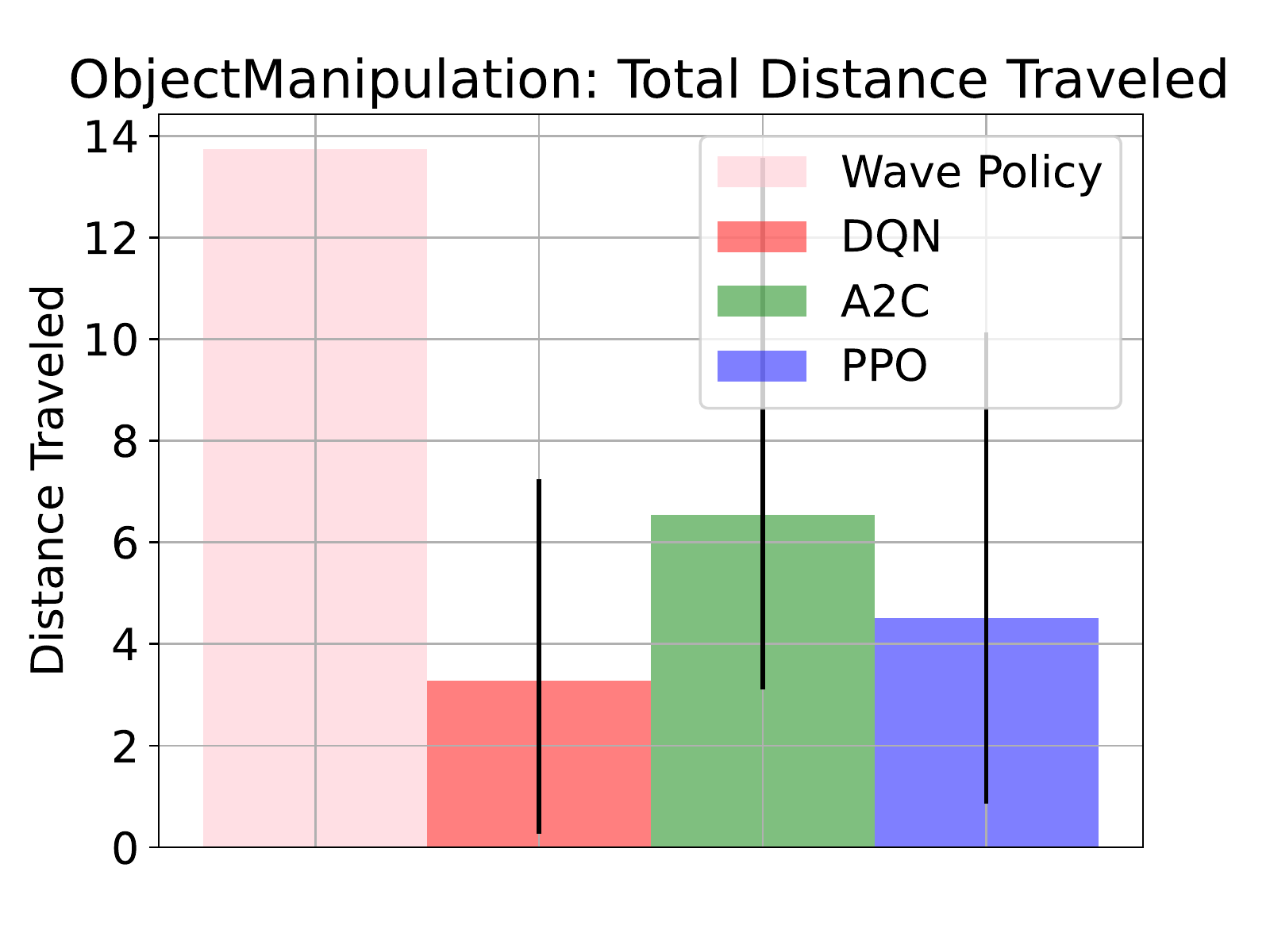}\\[\abovecaptionskip]
     \end{tabular}
     
     \begin{tabular}{@{}c@{}c@{}c@{}}
         \includegraphics[scale=0.25]{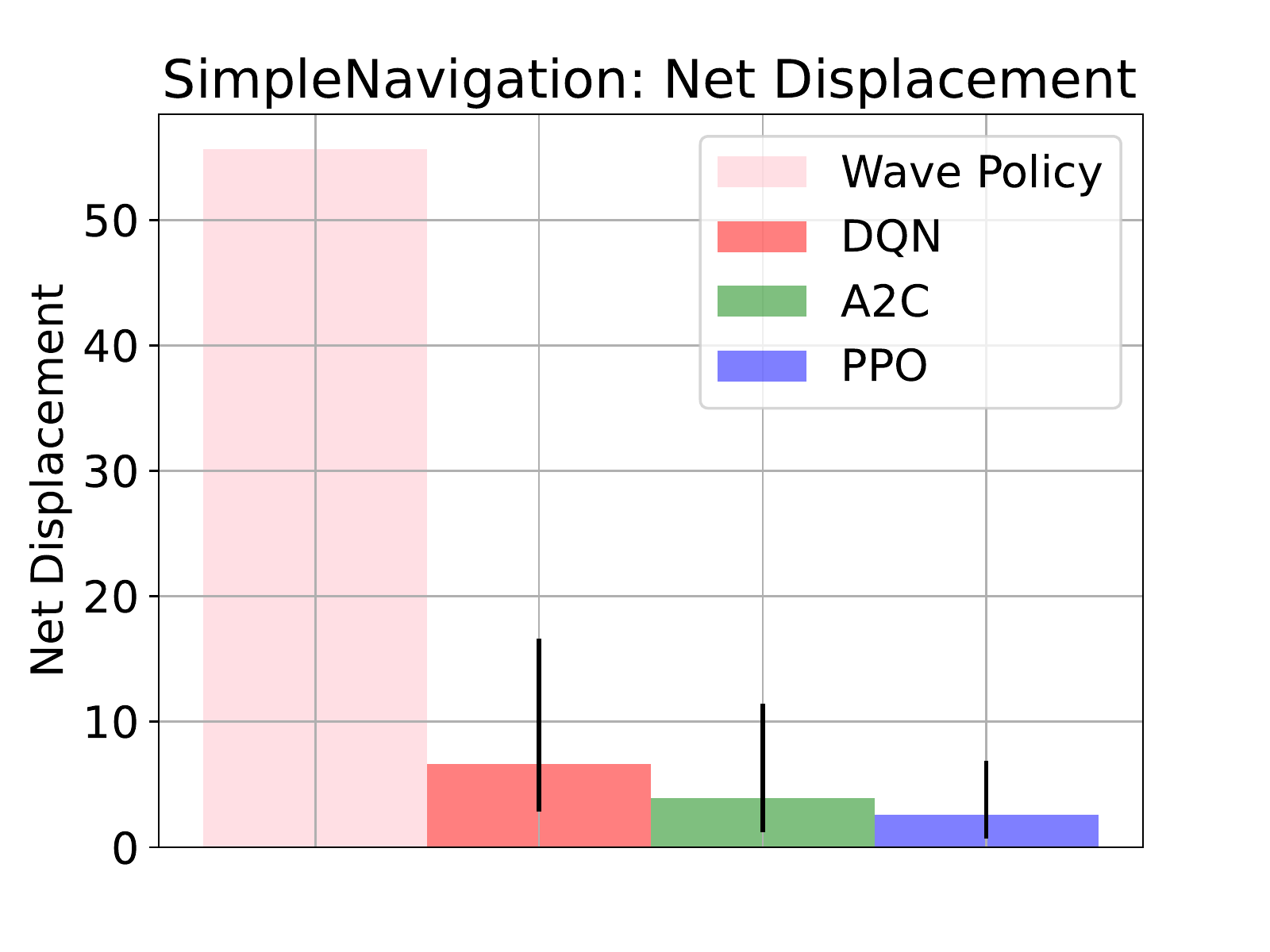}\\[\abovecaptionskip]
     \end{tabular}
     \begin{tabular}{@{}c@{}c@{}c@{}}
         \includegraphics[scale=0.25]{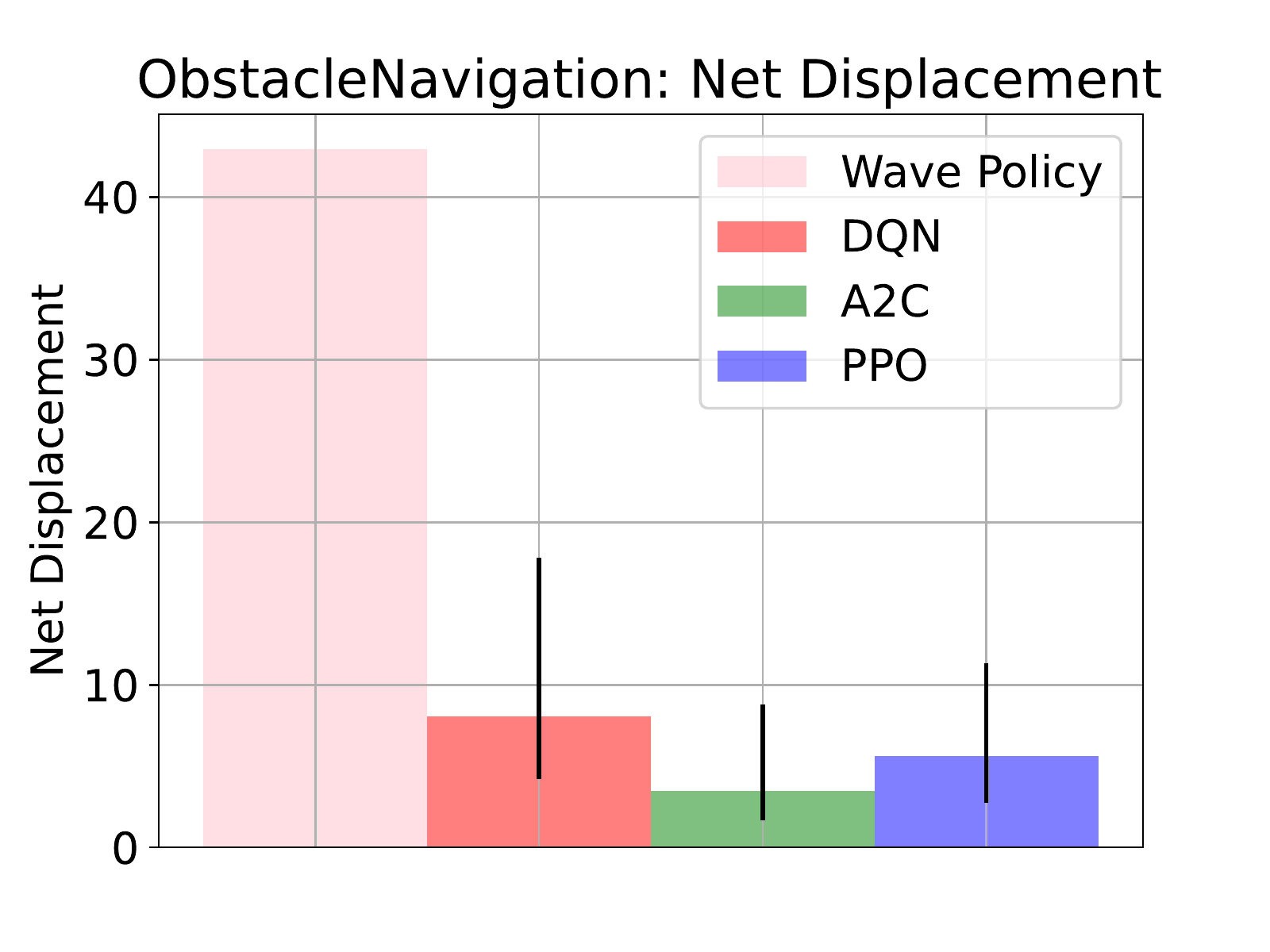}\\[\abovecaptionskip]
     \end{tabular}
     \begin{tabular}{@{}c@{}c@{}c@{}}
         \includegraphics[scale=0.25]{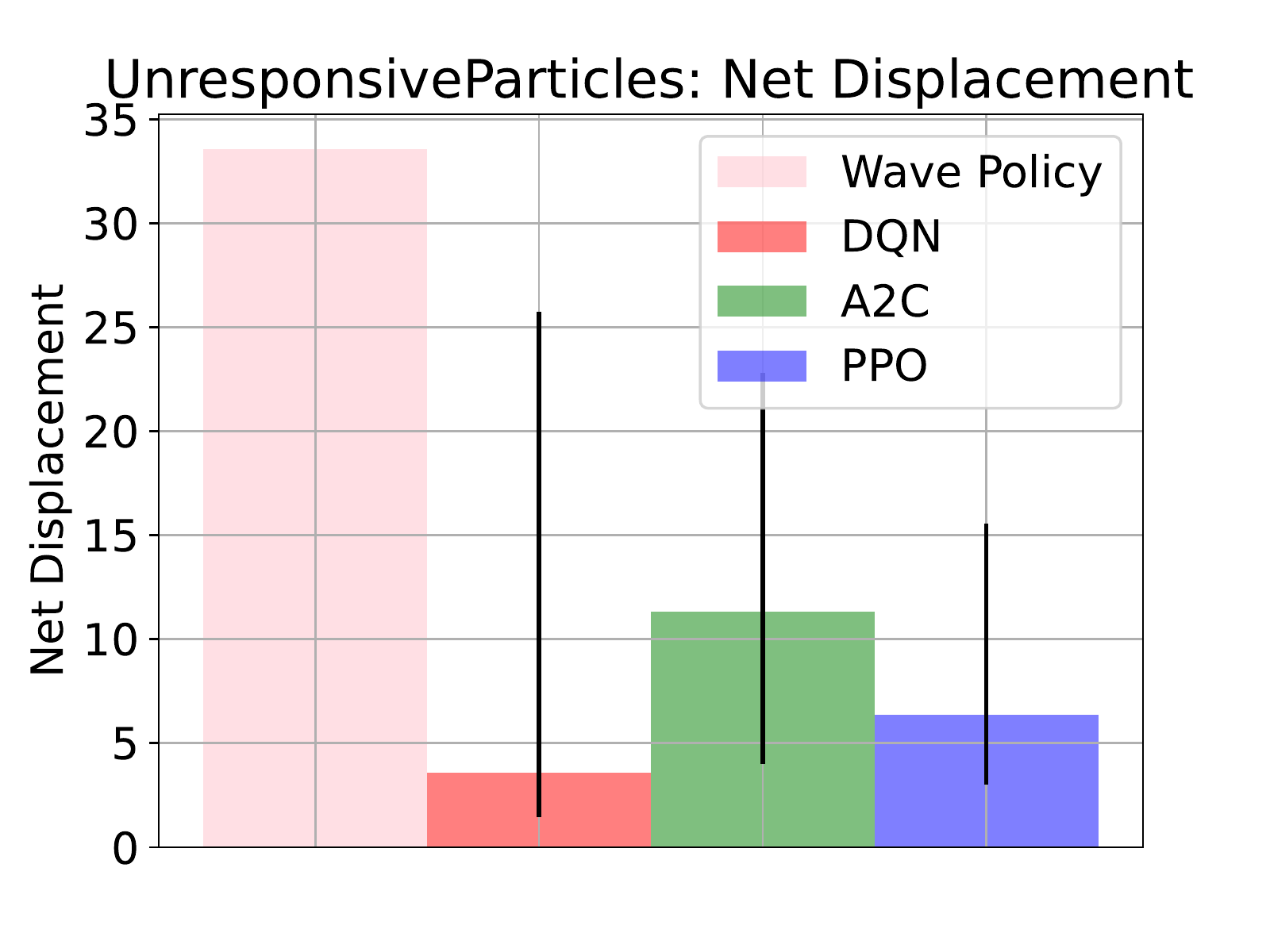}\\[\abovecaptionskip]
     \end{tabular}
     \begin{tabular}{@{}c@{}c@{}c@{}}
         \includegraphics[scale=0.25]{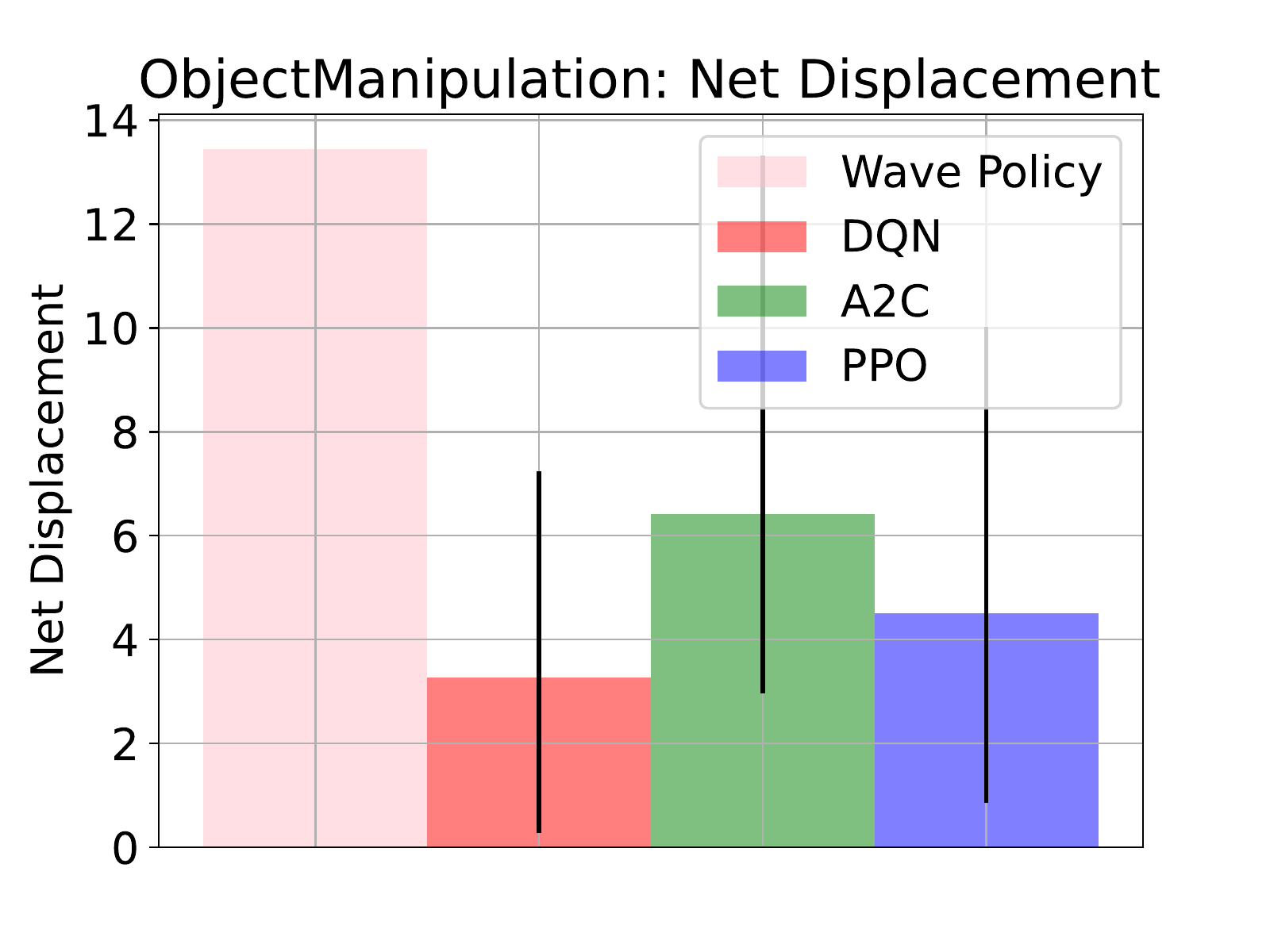}\\[\abovecaptionskip]
     \end{tabular}
     
     \begin{tabular}{@{}c@{}c@{}c@{}}
         \includegraphics[scale=0.25]{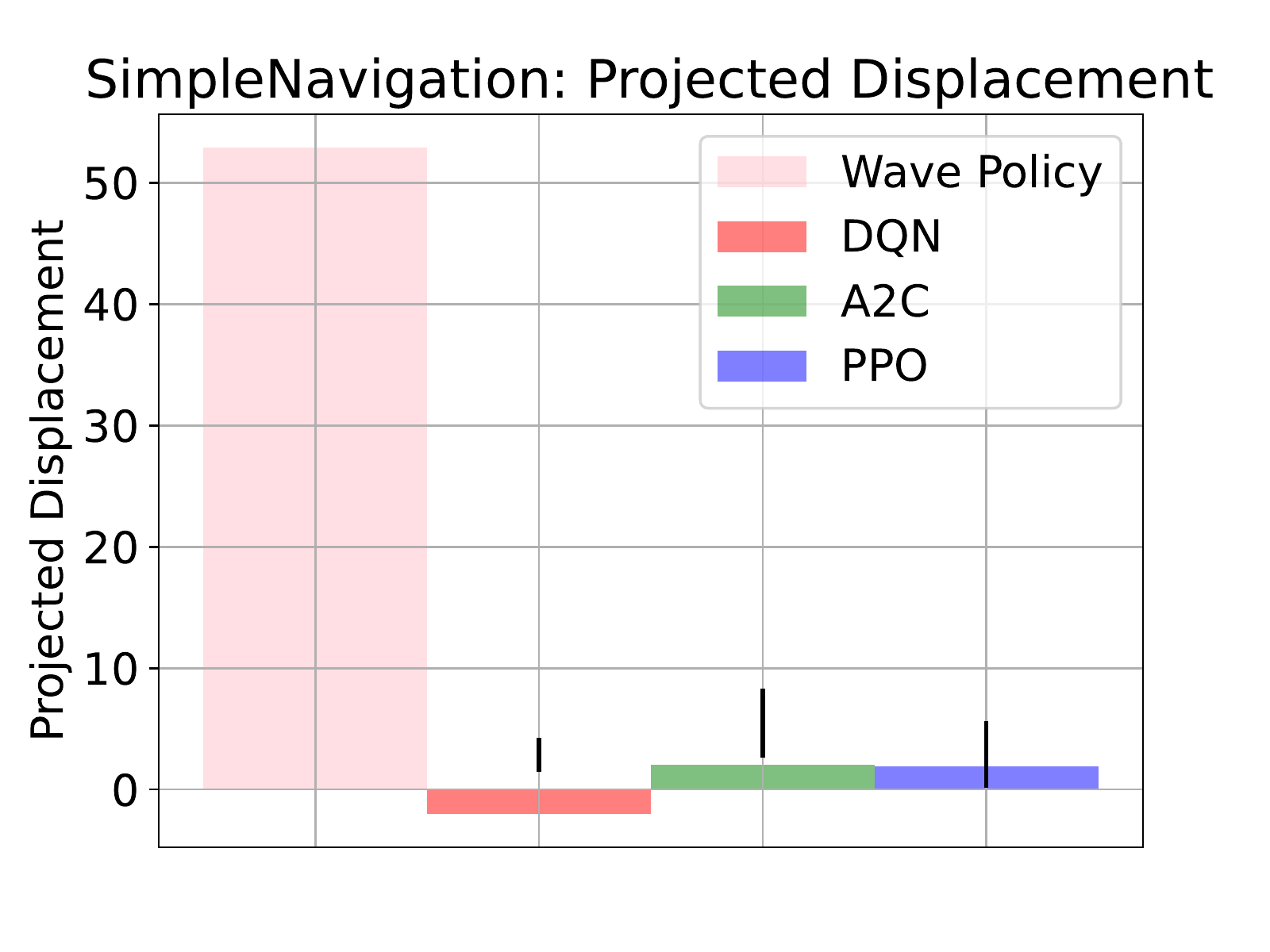}\\[\abovecaptionskip]
     \end{tabular}
     \begin{tabular}{@{}c@{}c@{}c@{}}
         \includegraphics[scale=0.25]{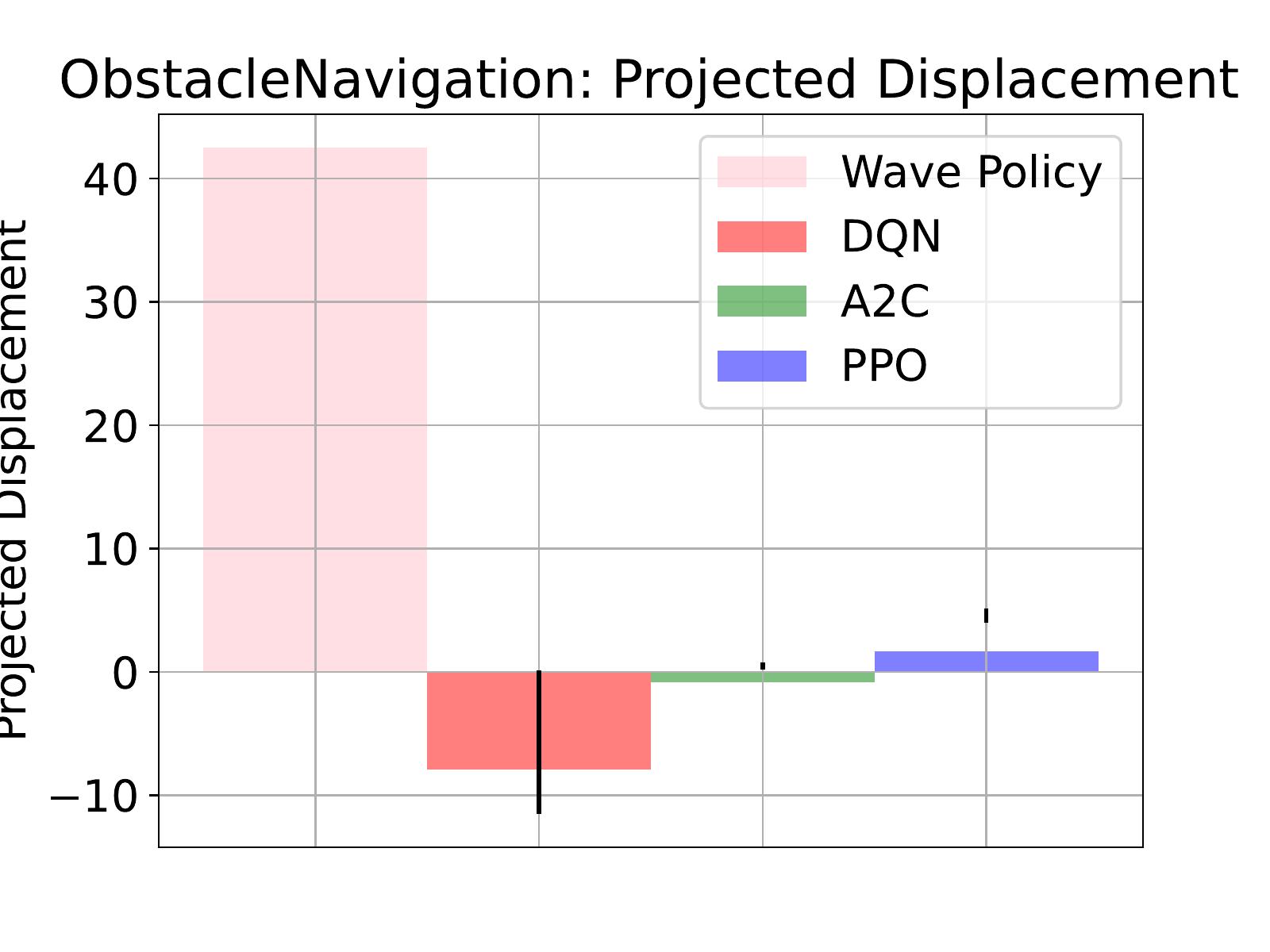}\\[\abovecaptionskip]
     \end{tabular}
     \begin{tabular}{@{}c@{}c@{}c@{}}
         \includegraphics[scale=0.25]{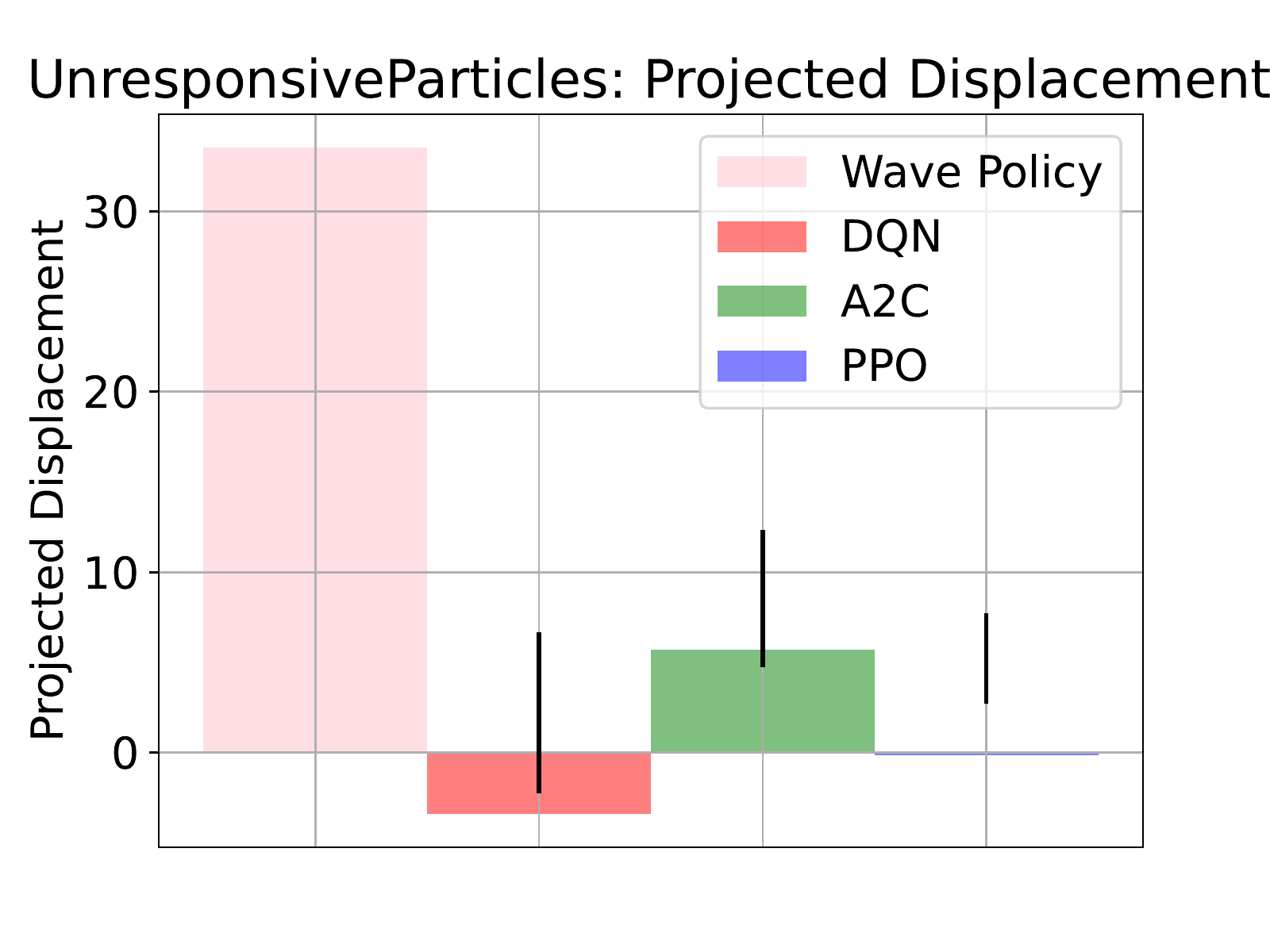}\\[\abovecaptionskip]
     \end{tabular}
     \begin{tabular}{@{}c@{}c@{}c@{}}
         \includegraphics[scale=0.25]{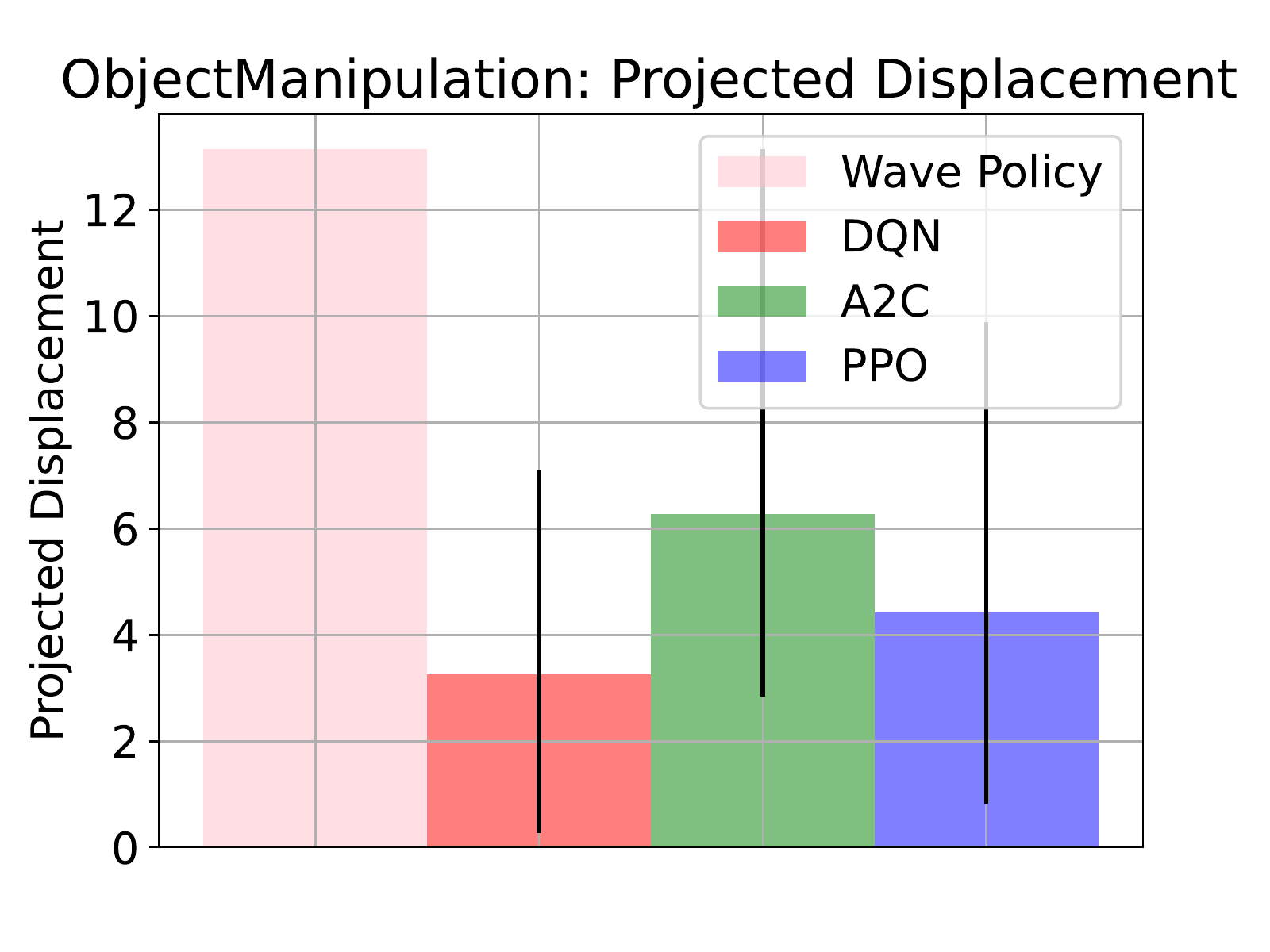}\\[\abovecaptionskip]
     \end{tabular}
     
    \caption{Benchmark results for all baseline models (handcrafted baseline, DQN, A2C, PPO): median performance plotted in bar graphs and maximum and minimum performance plotted in error bars.}
    \label{fig:performance bars}
\end{figure*}

\textbf{Navigation tasks}. From Figure \ref{fig:performance bars}, the DRL algorithms are outperformed drastically by the handcrafted algorithm. The Projected Displacement shows that in eight out of nine navigation tasks, the particle robots move less than one unit in the correct direction. However in Row 2, we see that the net displacement of the particle robots in the simple navigation task for DQN, A2C, and PPO is a significant amount, 10.8\%, 9.1\% , and 6.6\% of the handcrafted algorithm respectively. Figure \ref{fig:p2p trajectory} qualitatively corroborates these findings; the particle robot systems move significantly but in the incorrect direction during trial runs. From Row 1, we also notice that the total distance traveled by the handcrafted algorithm is nearly identical to the DRL algorithms. This suggests that existing DRL algorithms are beginning to solve a self-propulsion problem but mainly struggle with navigation. 

The handcrafted algorithm's performance significantly decreases as the tasks become more challenging. This is evident in both columns 2 and 3, where the handcrafted algorithm is ~75\% as effective in obstacle navigation and ~56\% as effective with unresponsive particle robots. However DRL algorithms do not have this clear drop-off with harder tasks. Using the net displacement metric, DQN and PPO performed better in the obstacle navigation than the simple navigation task. A2C and PPO performed better with unresponsive particles. PPO's net displacement performance relative to the handcrafted algorithm increased from 9.1\% in the simple navigation task to 25.4\% in the obstacle navigation task and 26.5\% with unresponsive particles. In potential real-world particle robot systems, DRL algorithms posses a greater ability to adapt to unforeseen challenges than the handcrafted algorithm. 

\textbf{Manipulation tasks}.
From the last column of Figure~\ref{fig:performance bars}, we notice that DQN and PPO are 9\% as effective as the handcrafted algorithm, and A2C is 34\% as effective. A2C was the best performing DRL model in the manipulation task. For each policy, the evaluation metrics: total distance, net displacement, and projected displacement, are all similar to each other. The particle robots are only able to push the target object, meaning any oscillatory motion by the particle robots is not reflected in the target object's motion. In contrast to the navigation tasks, the particle robots are able to move and navigate the object (\(\frac{\text{net displacement}}{\text{projected displacement}} > 0.97\) for all DRL models, meaning the target object is being pushed mostly in the correct direction), but DRL still lags behind handcrafted control.

\section{Conclusions}
Current deep reinforcement learning algorithms are far from matching the performance of a handcrafted control policy. These DRL algorithms struggle with navigation but possess limited locomotive abilities. The most encouraging result from the DRL algorithms was the performance in more challenging settings was similar to the simple navigation task, indicating a high level of adaptability. 

A potential reason for the unsatisfying performance of the particle robots is that the probability of the super-agent guessing a good action during training is very low. Therefore, the training is heavily dependent on the random search of the super-agent's action space, which is inefficient. In the future, we will study imitation learning on the handcrafted wave policy and examine supervised learning approach.

We hope to stimulate research into deep reinforcement learning for swarm robots by proposing a suite of navigation and construction tasks and benchmarking the performance of three state-of-the-art DRL algorithms and a handcrafted policy. We plan on introducing new particle robot designs with similar functions to study how mixed particle robot systems operate. We also plan to expand our environment into 3d and improve robot realism. Although the super-agent assumption is still a challenging task, we plan on benchmarking the environment in a decentralized multi-agent setting in order to fully extract the advantages of a low-level swarm robot system.

\textbf{Acknowledgment.}
This research is supported by the NSF CPS program under CMMI-1932187.

\addtolength{\textheight}{-0.2cm}   


\bibliographystyle{IEEEtranN}
\bibliography{ai4ce-tpl}

\begin{thebibliography}{28}
\providecommand{\natexlab}[1]{#1}
\providecommand{\url}[1]{#1}
\csname url@samestyle\endcsname
\providecommand{\newblock}{\relax}
\providecommand{\bibinfo}[2]{#2}
\providecommand{\BIBentrySTDinterwordspacing}{\spaceskip=0pt\relax}
\providecommand{\BIBentryALTinterwordstretchfactor}{4}
\providecommand{\BIBentryALTinterwordspacing}{\spaceskip=\fontdimen2\font plus
\BIBentryALTinterwordstretchfactor\fontdimen3\font minus
  \fontdimen4\font\relax}
\providecommand{\BIBforeignlanguage}[2]{{%
\expandafter\ifx\csname l@#1\endcsname\relax
\typeout{** WARNING: IEEEtranN.bst: No hyphenation pattern has been}%
\typeout{** loaded for the language `#1'. Using the pattern for}%
\typeout{** the default language instead.}%
\else
\language=\csname l@#1\endcsname
\fi
#2}}
\providecommand{\BIBdecl}{\relax}
\BIBdecl

\bibitem[Mateos(2020)]{mateos2020particle}
L.~A. Mateos, ``Particle robots a new species of hybrid bio-inspired
  robotics,'' \emph{arXiv:2003.08289}, 2020.

\bibitem[Li et~al.(2019)Li, Batra, Brown, Chang, Ranganathan, Hoberman, and
  Lipson]{shuguang2019paricle}
S.~Li, R.~Batra, D.~Brown, H.-D. Chang, N.~Ranganathan, C.~Hoberman, and D.~R.
  .~H. Lipson, ``Particle robotics based on statistical mechanics of loosely
  coupled components,'' \emph{Nature 567, 361–365}, 2019.

\bibitem[Mnih et~al.(2015)Mnih, Kavukcuoglu, Silver, Rusu, Veness, Bellemare,
  Graves, Riedmiller, Fidjeland, Ostrovski, Petersen, Beattie, Sadik,
  Antonoglou, King, Kumaran, Wierstra, and Hassabis]{volodymyr2019dqn}
V.~Mnih, K.~Kavukcuoglu, D.~Silver, A.~A. Rusu, J.~Veness, M.~G. Bellemare,
  A.~Graves, M.~Riedmiller, A.~K. Fidjeland, G.~Ostrovski, S.~Petersen,
  C.~Beattie, A.~Sadik, I.~Antonoglou, H.~King, D.~Kumaran, D.~Wierstra, and
  S.~L. .~D. Hassabis, ``Human-level control through deep reinforcement
  learning,'' \emph{Nature 518, 529–533}, 2015.

\bibitem[Mnih et~al.(2016)Mnih, Badia1, Mirza, Graves, Harley, Lillicrap,
  Silver, and Kavukcuoglu]{mnih2016a2c}
V.~Mnih, A.~P. Badia1, M.~Mirza, A.~Graves, T.~Harley, T.~P. Lillicrap,
  D.~Silver, and K.~Kavukcuoglu, ``Asynchronous methods for deep reinforcement
  learning,'' \emph{arXiv:1602.01783v2}, 2016.

\bibitem[Schulman et~al.(2017)Schulman, Wolski, Dhariwal, Radford, and
  Klimov]{schulman2017ppo}
J.~Schulman, F.~Wolski, P.~Dhariwal, A.~Radford, and O.~Klimov, ``Proximal
  policy optimization algorithms,'' \emph{arXiv:1707.06347v2}, 2017.

\bibitem[Savoie et~al.(2019)Savoie, Berrueta, Jackson, Pervan, Warkentin, Li,
  Murphey, and Goldman]{savoie2019smarticle}
W.~Savoie, T.~A. Berrueta, Z.~Jackson, A.~Pervan, R.~Warkentin, S.~Li, T.~D.
  Murphey, and K.~W. . D.~I. Goldman, ``A robot made of robots: Emergent
  transport and control of a smarticle ensemble,'' \emph{Science Robotics,
  Volume 4, Issue 34}, 2019.

\bibitem[Chvykov et~al.(2021)Chvykov, Berrueta, Vardhan, Savoie, Samland,
  Murphey, Wiesenfeld, Goldman, L., and England]{chvykov2021rattling}
P.~Chvykov, T.~A. Berrueta, A.~Vardhan, W.~Savoie, A.~Samland, T.~D. Murphey,
  K.~Wiesenfeld, D.~I. Goldman, J.~L., and England, ``Low rattling: A
  predictive principle for self-organization in active collectives,''
  \emph{Science, Vol 371, Issue 6524 • pp. 90-95}, 2021.

\bibitem[Deblais et~al.(2018)Deblais, Barois, Guérin, Delville, Vaudaine,
  Lintuvuori, Boudet, Baret, and Kellay]{deblais2018boundaries}
A.~Deblais, T.~Barois, T.~Guérin, P.-H. Delville, R.~Vaudaine, J.~S.
  Lintuvuori, J.-F. Boudet, J.-C. Baret, and H.~Kellay, ``Boundaries control
  collective dynamics of inertial self-propelled robots,'' \emph{Physical
  Review Letters, 120 (18), pp.188002 (1-5)}, 2018.

\bibitem[Boudet et~al.(2021)Boudet, Lintuvuori, Lacouture, Barois, Deblais,
  Xie, Cassagnere, Tregon, Bruckner, Baret, and Kellay]{boudet2021scaffold}
J.~F. Boudet, J.~Lintuvuori, C.~Lacouture, T.~Barois, A.~Deblais, K.~Xie,
  S.~Cassagnere, B.~Tregon, D.~B. Bruckner, J.~C. Baret, and H.~Kellay, ``From
  collections of independent, mindless robots to flexible, mobile and
  directional superstructures,'' \emph{SCIENCE ROBOTICS Vol 6, Issue 56}, 2021.

\bibitem[Wang et~al.(2020)Wang, Wang, Wang, Dong, and Wang]{wang2020distribute}
J.~Wang, T.~Wang, W.~Wang, X.~Dong, and Y.~Wang, ``Distributed localization
  without direct communication inspired by statistical mechanics,''
  \emph{arXiv:2006.02658v1}, 2020.

\bibitem[Mayya et~al.(2019)Mayya, Notomista, Shell, Hutchinson, and
  Egerstedt]{8967985}
S.~Mayya, G.~Notomista, D.~Shell, S.~Hutchinson, and M.~Egerstedt,
  ``Non-uniform robot densities in vibration driven swarms using phase
  separation theory,'' in \emph{2019 IEEE/RSJ International Conference on
  Intelligent Robots and Systems (IROS)}, 2019, pp. 4106--4112.

\bibitem[Yang et~al.(2020)Yang, Yu, and Zhang]{8884178}
L.~Yang, J.~Yu, and L.~Zhang, ``Statistics-based automated control for a swarm
  of paramagnetic nanoparticles in 2-d space,'' \emph{IEEE Transactions on
  Robotics}, vol.~36, no.~1, pp. 254--270, 2020.

\bibitem[Fan et~al.(2018)Fan, Long, Liu, and Pan]{fan2018collision}
T.~Fan, P.~Long, W.~Liu, and J.~Pan, ``Fully distributed multi-robot collision
  avoidance via deep reinforcement learning for safe and efficient navigation
  in complex scenarios,'' \emph{arXiv:1808.03841v1 [cs.RO] 11 Aug 2018}, 2018.

\bibitem[Long et~al.(2018)Long, Fan, Liao, Liu, Zhang, and
  Pan]{long2018collision}
P.~Long, T.~Fan, X.~Liao, W.~Liu, H.~Zhang, and J.~Pan, ``Towards optimally
  decentralized multi-robot collision avoidance via deep reinforcement
  learning,'' \emph{International Conference on Robotics and Automation (ICRA),
  pp. 6252-6259}, 2018.

\bibitem[Arya~Kumar(2019)]{rldrone2016arya}
A.~Z. Arya~Kumar, ``Using reinforcement learning for real-time trajectory
  planning of aerial multi agent systems,'' \emph{TJHSST Computer Systems
  Research Lab, Senior Research}, 2019.

\bibitem[Wang et~al.(2019)Wang, Cao, Stojmenovic, Zhao, Chen, and
  Jiang]{wang2019pattern}
J.~Wang, J.~Cao, M.~Stojmenovic, M.~Zhao, J.~Chen, and S.~Jiang, ``Pattern-rl:
  Multi-robot cooperative pattern formation via deep reinforcement learning,''
  \emph{International Conference on Machine Learning and Applications (ICMLA)},
  2019.

\bibitem[Ding et~al.(2020)Ding, Koh, Merckaert, Vanderborght, Nicotra, Heckman,
  Roncone, and Chen]{ding2020distributed}
G.~Ding, J.~J. Koh, K.~Merckaert, B.~Vanderborght, M.~M. Nicotra, C.~Heckman,
  A.~Roncone, and L.~Chen, ``Distributed reinforcement learning for cooperative
  multi-robot object manipulation,'' \emph{arXiv:2003.09540}, 2020.

\bibitem[ZHU et~al.(2020)ZHU, DAI, YAO, MA, ZENG, and LU]{zhu2020flocking}
P.~ZHU, W.~DAI, W.~YAO, J.~MA, Z.~ZENG, and H.~LU, ``Multi-robot flocking
  control based on deep reinforcement learning,'' \emph{IEEE Access, vol. 8,
  pp. 150397-150406, 2020}, 2020.

\bibitem[SUN et~al.(2021)SUN, ZHANG, YU, and ZHANG]{sun2021motion}
H.~SUN, W.~ZHANG, R.~YU, and Y.~ZHANG, ``Motion planning for mobile
  robots—focusing on deep reinforcement learning: A systematic review,''
  \emph{IEEE Access, vol. 9, pp. 69061-69081, 2021}, 2021.

\bibitem[Jiang(2018)]{jiang2018}
S.~Jiang, ``Multi-agent-reinforcement-learning-environment,''
  \url{https://github.com/Bigpig4396/Multi-Agent-Reinforcement-Learning-Environment},
  2018.

\bibitem[Zheng et~al.(2017)Zheng, Yang, Cai, Zhang, Wang, and
  Yu]{zheng2017magent}
L.~Zheng, J.~Yang, H.~Cai, W.~Zhang, J.~Wang, and Y.~Yu, ``Magent: A many-agent
  reinforcement learning platform for artificial collective intelligence,''
  \emph{arXiv:1712.00600}, 2017.

\bibitem[Lowe et~al.(2020)Lowe, Wu, Tamar, Harb, Abbeel, and
  Mordatch]{Lowe2020maparicle}
R.~Lowe, Y.~Wu, A.~Tamar, J.~Harb, P.~Abbeel, and I.~Mordatch, ``Multi-agent
  actor-critic for mixed cooperative-competitive environments,''
  \emph{arXiv:1706.02275v4}, 2020.

\bibitem[Baker et~al.(2020)Baker, Kanitscheider, Markov, Wu, Powell, McGrew,
  and Mordatch]{Baker2020maEmergent}
B.~Baker, I.~Kanitscheider, T.~Markov, Y.~Wu, G.~Powell, B.~McGrew, and
  I.~Mordatch, ``Emergent tool use from multi-agent autocurricula,''
  \emph{arXiv:1909.07528}, 2020.

\bibitem[pla(2019)]{playground}
``Playground,'' \url{https://github.com/MultiAgentLearning/playground}, 2019.

\bibitem[Suarez(2019)]{neural-mmo}
J.~Suarez, ``Neural mmos,'' \url{https://github.com/openai/neural-mmo}, 2019.

\bibitem[Samvelyan(2019)]{starcraft}
M.~Samvelyan, ``Starcraft ii,'' \url{https://github.com/oxwhirl/smac}, 2019.

\bibitem[foo(2019)]{football}
``Google research football,''
  \url{https://github.com/google-research/football}, 2019.

\bibitem[Raffin et~al.(2019)Raffin, Hill, Ernestus, Gleave, Kanervisto, and
  Dormann]{stable-baselines3}
A.~Raffin, A.~Hill, M.~Ernestus, A.~Gleave, A.~Kanervisto, and N.~Dormann,
  ``Stable baselines3,'' \url{https://github.com/DLR-RM/stable-baselines3},
  2019.

\end{thebibliography}

\end{document}